\documentclass[lettersize,journal]{IEEEtran}
\usepackage{amsmath,amsfonts}
\usepackage{algorithmic}
\usepackage{algorithm}
\usepackage{array}
\usepackage[caption=false,font=normalsize,labelfont=sf,textfont=sf]{subfig}
\usepackage{textcomp}
\usepackage{stfloats}
\usepackage{url}
\usepackage{verbatim}
\usepackage{graphicx}
\usepackage{cite}
\usepackage{color}
\usepackage{mathtools}
\usepackage{multirow}
\usepackage{tipa}
\usepackage{xspace}
\usepackage{bm} 

\makeatletter
\DeclareRobustCommand\onedot{\futurelet\@let@token\@onedot}
\def\@onedot{\ifx\@let@token.\else.\null\fi\xspace}

\def\eg{\emph{e.g}\onedot} 
\def\ie{\emph{i.e}\onedot}

\makeatother

\newcommand{\hide}[1]{}

\newcommand\wh[1]{\hstretch{2}{\hat{\hstretch{.5}{#1\mkern1mu}}}\mkern-1mu}
\usepackage{color}
\usepackage{xcolor,colortbl}
\usepackage{booktabs}
\usepackage{mathtools}
\usepackage{multirow}
\usepackage{tipa}
\usepackage{scalerel}

\newcommand{\tit}[1]{\smallbreak\noindent\textbf{#1}}

\definecolor{purple}{rgb}{0.65,0,0.65}
\definecolor{dark_green}{rgb}{0, 0.5, 0}
\definecolor{blueish}{rgb}{0.0, 0.3, .6}
\definecolor{LightCyan}{rgb}{0.88,0.95,1}

\definecolor{tabhighlight}{HTML}{e5e5e5}
\newcolumntype{h}{>{\columncolor{tabhighlight}}c}

\newcommand{\methodiccv}{SEARLE\xspace}
\newcommand{\methodpami}{iSEARLE\xspace}
\newcommand{\RNum}[1]{\lowercase\expandafter{\romannumeral #1\relax}}

\newcommand{\extmethod}{improved zero-Shot composEd imAge Retrieval with textuaL invErsion\xspace}
\usepackage[pagebackref,breaklinks]{hyperref}
\DeclareRobustCommand{\vect}[1]{\bm{#1}}
\usepackage[capitalize]{cleveref}
\crefname{section}{Sec.}{Secs.}
\Crefname{section}{Section}{Sections}
\Crefname{table}{Table}{Tables}
\crefname{table}{Tab.}{Tabs.}

\begin{document}

\title{\methodpami: Improving Textual Inversion for\\Zero-Shot Composed Image Retrieval}

\author{Lorenzo Agnolucci$^*$, Alberto Baldrati$^*$, Alberto Del Bimbo, Marco Bertini
\thanks{* The first two authors contributed equally to this work.}
\thanks{L. Agnolucci, A. Baldrati, A. Del Bimbo, M. Bertini are with the Media Integration and Communication Center (MICC), University of Florence, Italy  (e-mail: [name].[surname]@unifi.it).}
}

\markboth{IEEE Transactions on Pattern Analysis and Machine Intelligence}{} 



\maketitle

\begin{abstract} Given a query consisting of a reference image and a relative caption, Composed Image Retrieval (CIR) aims to retrieve target images visually similar to the reference one while incorporating the changes specified in the relative caption. The reliance of supervised methods on labor-intensive manually labeled datasets hinders their broad applicability to CIR. In this work, we introduce a new task, Zero-Shot CIR (ZS-CIR), that addresses CIR without the need for a labeled training dataset. We propose an approach, named \methodpami (\extmethod), that involves mapping the visual information of the reference image into a pseudo-word token in the CLIP token embedding space and combining it with the relative caption. To foster research on ZS-CIR, we present an open-domain benchmarking dataset named CIRCO (Composed Image Retrieval on Common Objects in context), the first CIR dataset where each query is labeled with multiple ground truths and a semantic categorization. The experimental results illustrate that \methodpami obtains state-of-the-art performance on three different CIR datasets -- FashionIQ, CIRR, and the proposed CIRCO -- and two additional evaluation settings, namely domain conversion and object composition.
The dataset, code, and model are publicly available at \small{\href{https://github.com/miccunifi/SEARLE}{\url{https://github.com/miccunifi/SEARLE}}}.
\end{abstract}


\section{Introduction}

\begin{figure}
    \centering
    \includegraphics[width=\columnwidth]{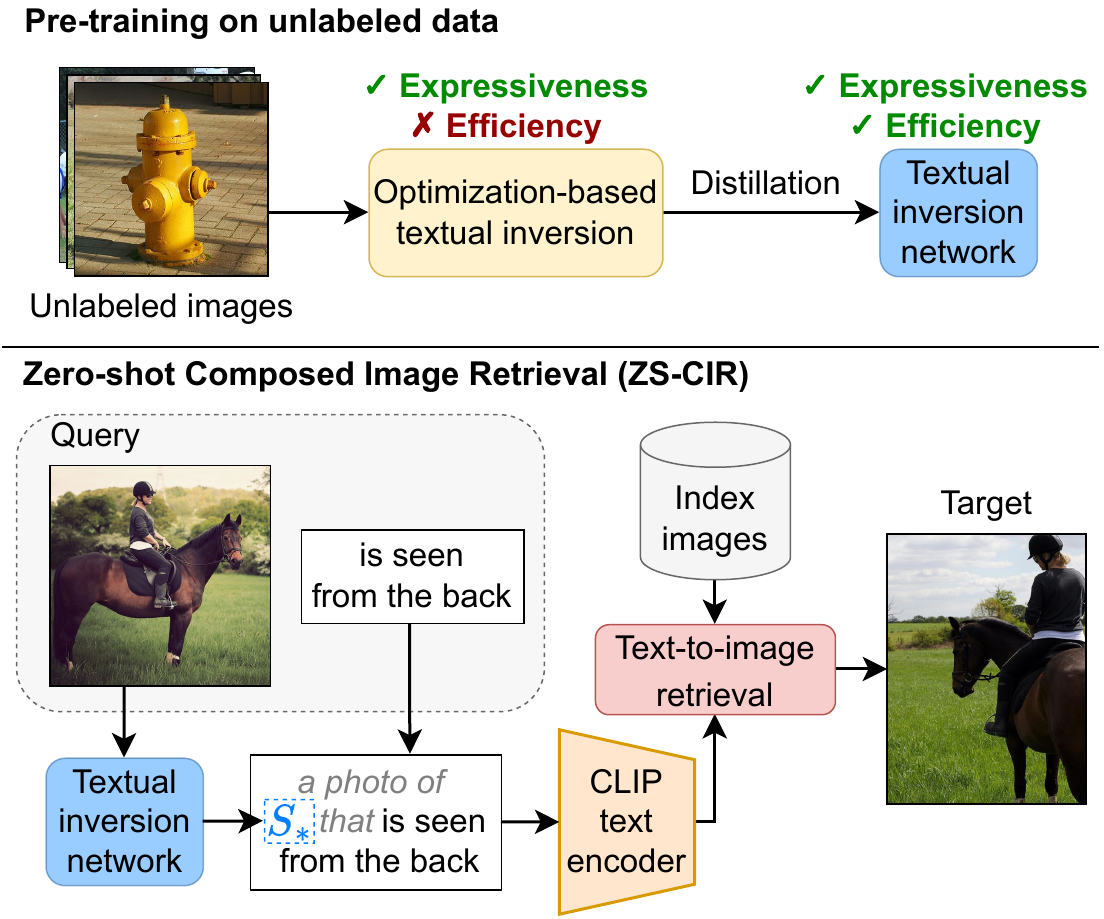}
      \caption{Workflow of our method. \textit{Top}: in the pre-training phase, we start by generating the pseudo-word tokens of unlabeled images with an expressive but computationally expensive optimization-based textual inversion. Then, we distill the knowledge embedded in the pseudo-word tokens into an expressive and efficient textual inversion network. \textit{Bottom}: at inference time on ZS-CIR, we use the textual inversion network to map the reference image to a pseudo-word $S_*$ and concatenate it with the relative caption. Then, we perform text-to-image retrieval using the features extracted with the CLIP text encoder.}
    \label{fig:intro}
\end{figure}

When provided with a query consisting of a reference image and a relative caption, Composed Image Retrieval (CIR) \cite{vo2019composing,liu2021image} seeks to retrieve target images that visually resemble the reference one while including the modifications described in the relative caption. The bi-modal structure of the query allows users to specify the desired image characteristics more precisely. It leverages the strengths of both language-based descriptions and visual features, as certain attributes are more effectively communicated through text, while others are better expressed visually. We provide some query examples in \cref{fig:circo_example}.

Datasets for CIR comprise triplets $(I_r, T_r, I_t)$, each including a reference image, a relative caption, and a target image, respectively. The creation of datasets for CIR is costly, primarily because such data is not readily accessible online, and automated generation remains a significant challenge. Consequently, researchers are compelled to undertake labor-intensive manual labeling efforts. The manual process entails the identification of reference and target image pairs and the composition of descriptive captions that outline the differences between them. Carrying out this task is both time-consuming and resource-intensive, particularly when building extensive training datasets.

Current works addressing CIR \cite{baldrati2022conditioned, baldrati2022effective,baldrati2023composed, wen2023target, delmasartemis, liu2021image, lee2021cosmo} depend on supervised learning to devise methods able to combine the reference image and the relative caption effectively. For example, \cite{baldrati2023composed} proposes a fully supervised two-stage approach based on fine-tuning CLIP encoders and training a combiner network. Despite their promising results, the dependence of current CIR approaches on expensive manually annotated datasets constrains their scalability and applicability in domains outside those of the training datasets.

Building on the conference version of this work \cite{baldrati2023zero}, we remove the need for costly labeled training data by introducing a new task: Zero-Shot Composed Image Retrieval (ZS-CIR). In ZS-CIR, the goal is to devise an approach capable of merging the information of the reference image and the relative caption without requiring supervised learning.

To address the challenges of ZS-CIR, we propose an approach named \methodpami \footnote{John Searle is an American philosopher renowned for his work on the philosophy of language and how words denote specific objects.} (\extmethod) based on the frozen pre-trained CLIP~\cite{radford2021learning} vision-language model. Our method simplifies CIR, casting it to a standard text-to-image retrieval task by mapping the reference image into a learned pseudo-word, which is subsequently appended to the relative caption. The pseudo-word corresponds to a pseudo-word token residing in the CLIP token embedding space. We refer to this mapping process with \textit{textual inversion}, following the terminology introduced in \cite{gal2022image}. \methodpami involves the pre-training of a textual inversion network -- denoted as $\phi$ -- on an unlabeled image-only dataset. The pre-training process consists of two stages: an Optimization-based Textual Inversion (OTI) with a GPT-powered regularization loss aimed at generating a set of pseudo-word tokens, and the distillation of their knowledge to $\phi$. Upon completion of the training, the network $\phi$ is capable of carrying out textual inversion in a single forward pass. During inference, when presented with a query $(I_r, T_r)$, we employ $\phi$ to predict the pseudo-word corresponding to $I_r$ and then concatenate it to $T_r$. Afterward, we exploit the CLIP common embedding space to perform text-to-image retrieval. \Cref{fig:intro} shows the workflow of the proposed approach.

The majority of existing CIR datasets focus on specific domains, such as fashion \cite{berg2010automatic, han2017automatic, wu2021fashion, guo2018dialog}, birds \cite{forbes2019neural}, or synthetic objects \cite{vo2019composing}. To the best of our knowledge, the CIRR dataset \cite{liu2021image} stands alone in encompassing natural images within an open domain. However, CIRR suffers from two main problems. Firstly, it includes numerous false negatives, potentially leading to imprecise performance evaluations. Secondly, the queries often neglect the visual content of the reference image, rendering the task addressable through standard text-to-image techniques, as shown by the results reported in \cref{tab:cirr_test}. Additionally, existing CIR datasets provide only a single annotated ground truth image per query. To address these shortcomings and foster research on ZS-CIR, we introduce an open-domain benchmarking dataset named CIRCO\footnote{CIRCO is pronounced as /\textipa{\textteshlig  \ignorespaces  irko}/.} (Composed Image Retrieval on Common Objects in context). CIRCO comprises validation and test sets derived from images within the COCO dataset~\cite{lin2014microsoft}. As a benchmarking dataset for ZS-CIR, a large training set is not needed, leading to a considerable reduction in labeling effort. To overcome the single ground truth limitation of existing CIR datasets, we propose to leverage our method to ease the annotation process of multiple ground truths. Consequently, CIRCO is the first CIR dataset with multiple annotated ground truths, enabling a more comprehensive evaluation of CIR models. In addition, contrary to existing CIR datasets, we provide a semantic categorization of the queries that allows a fine-grained semantic analysis of the results. We release only the validation set ground truths of CIRCO and host an evaluation server, enabling researchers to get performance metrics on the test set\footnote{Accessible at: \href{https://circo.micc.unifi.it/}{\url{https://circo.micc.unifi.it}}}.

The experimental results show that \methodpami achieves state-of-the-art performance on three different CIR datasets: FashionIQ \cite{wu2021fashion}, CIRR \cite{liu2021image}, and the proposed CIRCO. Moreover, the experiments on two additional settings, namely domain conversion and object composition \cite{saito2023pic2word}, prove that our model has better generalization capabilities than competing methods.

Our contributions can be summarized as follows:
\begin{itemize}
    \item We introduce a new task, Zero-Shot Composed Image Retrieval (ZS-CIR), to eliminate the requirement for costly labeled data for CIR;
    \item We propose a novel approach, named \methodpami, that relies on a textual inversion network to address ZS-CIR by mapping images into pseudo-words. Our method comprises two phases: an optimization-based textual inversion using a GPT-powered regularization loss and the training of the textual inversion network with a distillation loss;
    \item We introduce CIRCO, an open-domain benchmarking dataset for ZS-CIR with multiple annotated ground truths, reduced false negatives, and a semantic categorization of the queries. We propose to leverage our model to simplify the annotation process;
    \item \methodpami achieves state-of-the-art results on three different CIR datasets -- FashionIQ, CIRR, and the proposed CIRCO -- and two additional evaluation settings, \ie domain conversion and object composition.
\end{itemize}

This work extends our conference paper \cite{baldrati2023zero} in several aspects: 1) we improve our method by: i) adding Gaussian noise to the text features during OTI to mitigate the issue of the \textit{modality gap} \cite{liang2022mind}; ii) employing an additional regularization loss while training $\phi$ to prevent the predicted pseudo-word tokens from residing in sparse regions of the CLIP token embedding space; iii) proposing a hard negative sampling strategy to help $\phi$ in capturing fine-grained details; 2) we perform an additional annotation phase to allow a fine-grained semantic analysis on CIRCO, and we provide a more detailed study of our dataset; 3) we conduct more comprehensive experiments, including additional competitors and evaluation settings; 4) we perform a more thorough analysis of the proposed approach by studying the impact of the pre-training dataset and the effectiveness of the pseudo-word tokens in capturing visual information.

The remainder of this paper is organized as follows. \cref{sec:related_work} reviews related work. \cref{sec:proposed_approach} details our proposed approach.  \cref{sec:circo_dataset} describes the proposed CIRCO dataset. \cref{sec:experimental_results} presents experimental results and analysis. \cref{sec:conclusion} concludes the paper with final remarks.

\section{Related Work} \label{sec:related_work}

\tit{Composed Image Retrieval} CIR is a branch of compositional learning, an area that has been widely explored in various vision and language tasks. These include visual question answering~\cite{antol2015vqa, shao2023prompting}, image captioning~\cite{cornia2020meshed, barraco2023little}, and image synthesis~\cite{rombach2022high, meng2023distillation}. 
Compositional learning aims to create joint embedding features that effectively integrate and express information from both the textual and visual domains.

The research on CIR spans several domains, including fashion~\cite{berg2010automatic, han2017automatic, wu2021fashion, guo2018dialog}, natural images~\cite{forbes2019neural, liu2021image}, and synthetic images~\cite{vo2019composing}. 
The task was first introduced in \cite{vo2019composing}, where the authors propose a residual gating method for composing image-text features, aiming to merge multimodal information effectively. 
More recently, the use of the CLIP model as a backbone for CIR has received increasing attention \cite{baldrati2022effective, baldrati2023composed, baldrati2022conditioned, wen2023target}. \cite{baldrati2022effective} shows the effectiveness of combining out-of-the-box CLIP features with a Combiner network. Building on this, \cite{baldrati2023composed} introduces a task-specific fine-tuning step for CLIP encoders. Unlike the aforementioned approaches, the proposed method does not require supervision and uses unlabeled images for training, effectively learning to combine multimodal information without relying on a manually annotated CIR dataset.

\tit{Zero-Shot Composed Image Retrieval} The Zero-Shot Composed Image Retrieval (ZS-CIR) task was introduced concurrently by Pic2Word~\cite{saito2023pic2word} and the conference version of this work~\cite{baldrati2023zero}. Since its introduction, several works have proposed zero-shot approaches that do not rely on costly manually annotated datasets \cite{levy2023data, ventura2023covr, liu2023zero, chen2023pretrain, li2024catllm, karthik2024visionbylanguage, sun2023training, tang2023context, gu2023language}.

A line of research tackles ZS-CIR by substituting the manually labeled triplets with automatically constructed ones using an LLM~\cite{levy2023data, ventura2023covr, liu2023zero}. Specifically, \cite{levy2023data} proposes a \mbox{GPT-3}-based method \cite{brown2020language} for generating CIR triplets from an existing VQA dataset by leveraging question-answer pairs. A similar strategy is adopted by \cite{liu2023zero}, which uses ChatGPT to automatically construct the triplets starting from image-caption pairs. In contrast, \methodpami does not require any triplet-based training, as it relies only on unlabeled images. A different line of research also employs LLMs for ZS-CIR but uses them as auxiliary models at inference time rather than for automatic dataset construction~\cite{li2024catllm, karthik2024visionbylanguage, sun2023training}. For example, \cite{karthik2024visionbylanguage} presents a training-free approach that casts CIR to standard text-to-image retrieval by using an LLM to combine the relative caption with an automatically generated caption of the reference image. Despite the promising results, the reliance on an LLM at inference time introduces a non-negligible computational overhead when performing the retrieval. 

Among the approaches addressing ZS-CIR, the most similar to our work are \cite{tang2023context, gu2023language, saito2023pic2word}, as they present different methods for performing textual inversion while keeping the CLIP backbone frozen. 

\tit{Textual Inversion}\label{sec:related_work_textual_inversion}
In text-to-image synthesis, mapping images to a single pseudo-word is emerging as a powerful technique for generating personalized images~\cite{gal2022image, ruiz2023dreambooth, kumari2022multi}. \cite{gal2022image} performs textual inversion by relying on the reconstruction loss of a latent diffusion model~\cite{rombach2022high}. Additionally, \cite{ruiz2023dreambooth} also fine-tunes a pre-trained text-to-image diffusion model.

Besides personalized text-to-image synthesis, textual inversion has also been applied to image retrieval tasks~\cite{cohen2022this,saito2023pic2word, korbar2022personalised, gu2023language, tang2023context}. 
Specifically, PALAVRA~\cite{cohen2022this} addresses personalized image retrieval by pre-training a mapping function and then optimizing the predicted pseudo-word token at inference time. Several works employ textual inversion to address ZS-CIR \cite{saito2023pic2word, gu2023language, tang2023context}. LinCIR \cite{gu2023language} is a language-only approach for training the textual inversion network. Context-I2W \cite{tang2023context} is based on a transformer-based textual inversion network trained on the image-caption pairs of the CC3M dataset \cite{sharma2018conceptual}. Moreover, the textual inversion process is also dependent on the query text. This comes at the cost of requiring a double forward pass of the text encoder and a more complex network architecture than the proposed single MLP approach. The method most similar to ours is Pic2Word \cite{saito2023pic2word}. Pic2Word relies on a textual inversion network trained on the 3M images of CC3M using only a cycle contrastive loss. In contrast, we train our textual inversion network on only 3\% of the data and use a weighted sum of distillation and regularization losses. The distillation loss leverages the information provided by a set of pre-generated tokens obtained via optimization-based textual inversion.

\tit{Knowledge Distillation}
Knowledge distillation is a machine learning technique in which a simpler model (the student) learns to replicate the behavior of a more complex one (the teacher) by learning from its predictions \cite{hinton2015distilling}. This method has proven effective in various computer vision tasks, such as image classification \cite{hinton2015distilling, mistretta2024improving, beyer2022knowledge}, object detection \cite{chen2017learning, chawla2021data}, and text-to-image synthesis~\cite{meng2023distillation, sauer2023adversarial}, improving model compression, computational efficiency, and accuracy. In our work, we refer to knowledge distillation as the process of transferring knowledge from a computationally expensive optimization method (teacher) to a more efficient neural network (student). Specifically, we train a textual inversion network to emulate the output of an optimization-based textual inversion using a distillation loss. From a different point of view, our lightweight network can be viewed as a surrogate model of the more resource-intensive optimization technique.

\section{Proposed Approach} \label{sec:proposed_approach}

\begin{figure*}[!htb]
    \centering
    \includegraphics[width=\linewidth]{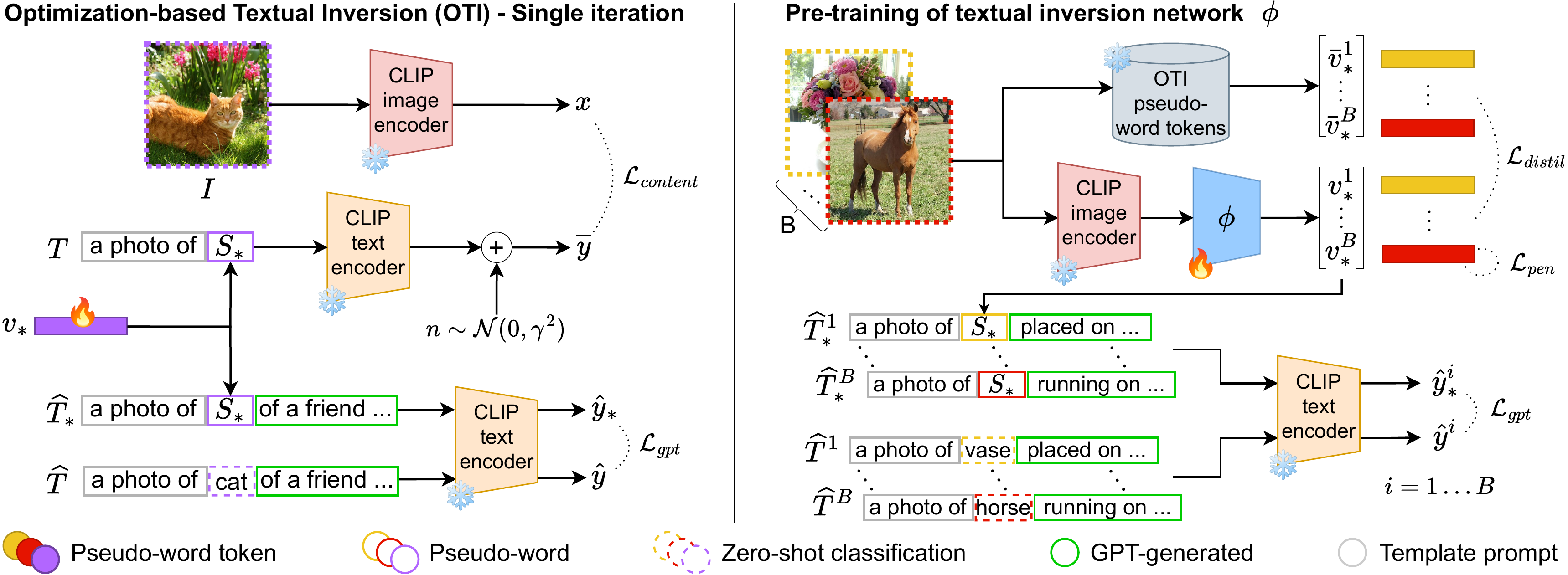}
    \vspace{-3.5ex}
    \caption{Overview of our approach. \textit{Left}: single iteration of the iterative Optimization-based Textual Inversion (OTI) method to generate a pseudo-word token $v_*$ from an image $I$. We force $v_*$ to represent the image content with a cosine loss $\mathcal{L}_{content}$. We add Gaussian noise to the text features before computing $\mathcal{L}_{content}$. We assign a concept word to $I$ with a CLIP zero-shot classification and feed the prompt ``a photo of \{concept\}" to GPT to continue the phrase, resulting in $\widehat{T}$. Let $S_*$ be the pseudo-word associated with $v_*$. We craft $\widehat{T}_*$ by replacing in $\widehat{T}$ the concept with $S_*$. $\widehat{T}$ and $\widehat{T}_*$ are then employed for a contextualized regularization with $\mathcal{L}_{gpt}$. \textit{Right}: pre-training of textual inversion network $\phi$ on unlabeled images. Given a set of pseudo-word tokens pre-generated with OTI, we distill their knowledge to $\phi$ through a contrastive loss $\mathcal{L}_{distil}$. We regularize the output of $\phi$ with the same GPT-powered loss $\mathcal{L}_{gpt}$ employed in OTI and an additional penalty term $\mathcal{L}_{pen}$. B represents the number of images in a batch.
    }
    \vspace{-4pt}
    \label{fig:teaser}
\end{figure*}

\tit{Preliminaries}
CLIP (Contrastive Language-Image Pre-training) \cite{radford2021learning} is a vision and language model trained on a large-scale dataset to align images and corresponding text captions in a common embedding space. CLIP is composed of an image encoder $\psi_{I}$ and a text encoder $\psi_{T}$. Given an image $I$, the image encoder extracts its feature representation $x = \psi_{I}(I) \in \mathbb{R}^{d}$, where $d$ is the size of CLIP embedding space. For a given text caption $T$, a word embedding layer $E_w$ maps each tokenized word to the token embedding space $\mathcal{W}$. Then, the text encoder $\psi_{T}$ generates the textual feature representation $y=\psi_{T}(E_w(T)) \in \mathbb{R}^{d}$ from the token embeddings. CLIP is trained to ensure that images and text expressing the same concepts correspond to similar feature representations within the shared embedding space.

\tit{Overview}
Starting from a frozen pre-trained CLIP model, the proposed method, named \methodpami, is designed to generate a representation of the reference image that can be used as input to the CLIP text encoder. We achieve this goal by mapping the visual features of the image into a new token embedding within the CLIP token embedding space $\mathcal{W}$. We term this token embedding \textit{pseudo-word token}, as it does not correspond to an actual word but rather serves as a representation of the image features within $\mathcal{W}$.

Our objective is dual. First, the pseudo-word token must accurately capture the content of the reference image. In other words, the text features related to a basic prompt comprising the pseudo-word should closely align with the corresponding image features. Second, the pseudo-word must effectively integrate and communicate with the text of the relative caption. While a single image can be mapped to multiple pseudo-word tokens, we opt for using a single one, as it proves to be sufficient to encode the information of an image effectively (see \cref{tab:vstar_image_retrieval}). Moreover, from our preliminary experiments, using a single pseudo-word token achieves better performance than relying on multiple ones, as also observed by \cite{gal2022image}.

\methodpami entails pre-training a textual inversion network $\phi$ using an unlabeled image-only dataset through a two-stage process. First, we rely on an Optimization-based Textual Inversion (OTI) method, which iteratively produces a set of pseudo-word tokens by exploiting a GPT-based regularization loss. Second, we train $\phi$ via knowledge distillation from the pre-generated pseudo-word tokens. The $\phi$ network outputs the pseudo-word token associated with an image in a single forward pass by taking as input its visual features, previously extracted via the CLIP image encoder.

At inference time, the input for CIR is given by a query $(I_r, T_r)$ corresponding respectively to the reference image and the relative caption. We generate the pseudo-word token $v_*$ associated with the reference image as $ v_* = \phi(I_r)$. We denote by $S_*$ the pseudo-word corresponding to the pseudo-word token $v_*$, \ie the counterpart of $v_*$ expressed in natural language. To effectively integrate the visual information of $I_r$ with $T_r$, we build the template ``a photo of $S_*$ that \textit{\{relative caption\}}" and extract its features via the CLIP text encoder. Note that these text features provide a multimodal representation of the reference image and its associated relative caption, as they encompass both textual and visual information. Finally, we carry out standard text-to-image retrieval within an image database using the extracted text features. We provide an overview of the workflow of our method in \cref{fig:intro}.

Fundamentally, both OTI and $\phi$ carry out the same operation, \ie mapping the visual features of an image into a pseudo-word token by means of a textual inversion. Consequently, one could directly utilize OTI at inference time without the need for $\phi$. However, $\phi$ offers considerably improved efficiency compared to OTI, which needs a non-negligible amount of time to be performed (see \cref{sec:implementation_details}). Considering that OTI has demonstrated its effectiveness in generating fruitful pseudo-word tokens (see \cref{sec:experimental_results}), we propose to distill their knowledge into a feed-forward network. Our approach strives to maintain the powerful expressiveness of OTI while achieving a negligible inference time. From now on, we refer to our approach as \methodpami when relying on $\phi$ for generating the pseudo-word token and as \methodpami-OTI when we directly utilize OTI for inference.

\subsection{Optimization-based Textual Inversion (OTI)}\label{sec:optimization}
Given an image $I$, we carry out textual inversion through an optimization-based approach that iteratively optimizes the pseudo-word token $v_* \in \mathcal{W}$ for a fixed number of iterations. The left section of \cref{fig:teaser} provides an overview of OTI.

First, we randomly initialize the pseudo-word token $v_*$ and associate the pseudo-word $S_*$ with it. Then, we craft a template sentence $T$, such as ``a photo of $S_*$", and process it with the CLIP text encoder $\psi_{T}$, resulting in $y = \psi_{T}(T)$. Following \cite{cohen2022this}, we randomly sample $T$ from a given set of templates. We employ the CLIP image encoder $\psi_{I}$ to extract the visual features $x = \psi_{I}(I)$.

Our goal is to obtain a pseudo-word token $v_*$ that captures the informative content of $I$. To this end, in our preliminary work \cite{baldrati2023zero}, we directly minimize the discrepancy between the image and text features by leveraging the CLIP common embedding space. However, \cite{liang2022mind} shows that in vision-language models such as CLIP the features associated with text and images correspond to different regions of the joint embedding space. In other words, text and image embeddings fall into separate clusters in the feature space. This phenomenon is commonly referred to as \textit{modality gap} \cite{liang2022mind, mistretta2025cross, gu2023can}. To mitigate this issue, \cite{gu2023can} proposes a simple and training-free strategy that involves adding Gaussian noise to the text features. Intuitively, the noise reduces the modality gap by spreading out the text embeddings to make them overlap with the image ones.

Inspired by \cite{gu2023can}, we propose to add Gaussian noise to the text features $y$ before minimizing their discrepancy with the image features $x$. Specifically, we compute $\overline{y} =  y + n$, where $n \sim \mathcal{N}(0, \gamma^2)$ is drawn from a Gaussian distribution with variance $\gamma^2$. Finally, we employ a cosine loss to maximize the similarity between the image and noisy text features:
\begin{equation}\label{eq:loss_content}
    \mathcal{L}_{content} = 1 - \cos{(x, \overline{y})}
\end{equation}
Therefore, differently from our preliminary work \cite{baldrati2023zero}, we mitigate the modality gap issue before computing the loss, resulting in improved performance (see \cref{sec:ablation_oti}). In addition, despite addressing the modality gap problem, \cite{gu2023can} does not directly contrast noisy text features with image ones, as the authors train their model without relying on visual data. On the contrary, we show that adding Gaussian noise to the text features is effective even when they are directly compared to the image features in the loss computation.

Relying solely on $\mathcal{L}_{content}$ is inadequate for generating a pseudo-word capable of interacting with other words of the CLIP dictionary. Indeed, similar to \cite{cohen2022this}, we observe that $\mathcal{L}_{content}$ pushes the pseudo-word token into sparse regions of CLIP token embedding space that differ from those encountered during CLIP's training. This phenomenon, akin to effects observed in GAN inversion works \cite{tov2021designing, zhu2020domain}, hampers the ability of the pseudo-word token to communicate effectively with other tokens. To address this limitation, we propose a novel regularization technique that constrains the pseudo-word token to reside on the CLIP token embedding manifold, thereby enhancing its interaction capabilities. Relying on CLIP zero-shot capabilities, we carry out a zero-shot classification of the image $I$. To classify the images, we employ a vocabulary originating from the $\sim$20K class names of the Open Images V7 dataset \cite{kuznetsova2020open}. Specifically, we assign the $k$ most similar distinct class names to each image, with $k$ being a hyperparameter. We will refer to these class names as \textit{concepts}, so, in other words, we associate each image to $k$ different concepts. Differently from \cite{cohen2022this}, we do not require the concepts as input.

After associating a set of concepts with an image, we generate a phrase using a lightweight GPT model \cite{brown2020language}. In each iteration of the optimization process, we randomly sample one of the $k$ concepts related to the image $I$ and feed the prompt ``a photo of \{concept\}" to GPT. Given that GPT is an autoregressive generative model, it manages to continue the prompt in a meaningful manner. For example, given the concept ```cat", the GPT-generated phrase might be $\wh{T}$ = ``a photo of cat that is eating in front of a window". In practice, since the vocabulary is known beforehand, we pre-generate all the GPT phrases for all the concepts in the vocabulary in advance. Starting from $\wh{T}$, we define $\wh{T}_*$ by simply replacing the concept with the pseudo-word $S_*$, resulting in $\wh{T}_*$ = ``a photo of $S_*$ that is eating\ldots". We extract the features of both phrases through the CLIP text encoder, obtaining  $\hat{y} = \psi_T(\wh{T})$ and $\hat{y}_* = \psi_T(\wh{T}_*)$. Finally, we rely on a cosine loss to maximize the similarity between the features:
\begin{equation}\label{eq:loss_oti_gpt}
    \mathcal{L}_{gpt} = 1 -\cos{(\hat{y}, \hat{y}_*)}
\end{equation}
Intuitively, $\mathcal{L}_{gpt}$ applies a contextualized regularization that steers $v_*$ toward the concept while considering a broader context. Indeed, compared to a generic pre-defined prompt, the GPT-generated phrases are more structured and thus similar to the relative captions used in CIR. In this way, we improve the ability of $v_*$ to interact with human-generated text such as the relative captions.

The final loss that we use for OTI is:
\begin{equation}\label{eq:loss_oti}
     \mathcal{L}_{OTI} =  \lambda_{content} \mathcal{L}_{content} + \lambda_{OTIgpt} \mathcal{L}_{gpt}
\end{equation}
where $\lambda_{content}$ and $\lambda_{OTIgpt}$ are the loss weights. Additionally, we find that applying a weight decay regularization to the pseudo-word tokens improves the effectiveness of the inversion process.

\subsection{Textual Inversion Network $\vect{\phi}$ Pre-training}\label{sec:phi_training}

OTI proves to be effective in generating pseudo-words that not only capture the visual information of an image but also interact fruitfully with actual words. However, its iterative and optimization-based nature results in a non-negligible amount of time for its execution (see \cref{sec:implementation_details}). To address this issue, we propose an approach for training a textual inversion network $\phi$ capable of predicting the pseudo-word tokens in a single forward pass by distilling knowledge from a collection of OTI pre-generated tokens. In other words, $\phi$ serves as a more efficient surrogate model of OTI, offering a faster and computationally less demanding approximation. The right part of \cref{fig:teaser} illustrates an overview of the pre-training phase.

We aim to obtain a single model capable of inverting images from any domain without the requirement of labeled training data. Specifically, we design an MLP-based textual inversion network $\phi$ with three linear layers, each followed by a GELU \cite{hendrycks2016gaussian} activation function and a dropout layer. 

Starting from an unlabeled pre-training dataset $\mathcal{D}$, we apply OTI to each image. Although this step is time-intensive, it is only required once, making it acceptable. This results in a collection of pseudo-word tokens, denoted as $\overline{\mathcal{V}}_* = \{{\bar{v}_*^j}\}_{j=1}^N$, where $N$ is the total number of images in $\mathcal{D}$. Our goal is to distill the knowledge captured by OTI in $\overline{\mathcal{V}}_*$ to $\phi$. Given an image $I \in \mathcal{D}$, we extract its features via the CLIP visual encoder, resulting in $x = \psi_{I}(I)$. We exploit $\phi$ to predict the pseudo-word token $v_* = \phi(x)$. We minimize the distance between the predicted pseudo-word token $v_*$ and the associated pre-generated token $\bar{v}_* \in \overline{\mathcal{V}}_*$ while maximizing the discriminability of each token. To achieve this, we employ a symmetric contrastive loss inspired by SimCLR \cite{chen2020simple, cohen2022this}:
\begin{align} \label{eq:loss_distil}
    \mathcal{L}_{distil} = &\frac{1}{B} \sum^{B}_{i=1} -\log{\frac{e^{( c(\bar{v}_*^i, v_*^i) / \tau )}}{\sum\limits^B_{j=1} {e^{( c(\bar{v}_*^i, v_*^j) / \tau )}}+ {\sum\limits_{j\neq i} {e^{( c(v_*^i, v_*^j) / \tau )}}} }}\nonumber \\ 
    &- \log{\frac{e^{( c(v_*^i, \bar{v}_*^i) / \tau )}}{\sum\limits^B_{j=1} {e^{( c(v_*^i, \bar{v}_*^j) / \tau )}}+ {\sum\limits_{j\neq i} {e^{( c(\bar{v}_*^i, \bar{v}_*^j) / \tau )}}} }} 
\end{align}
Here, $B$ is the number of images in a batch, $c(\cdot)$ indicates the cosine similarity, and $\tau$ is a temperature hyperparameter.

However, since the pre-training dataset $\mathcal{D}$ comprises real-world images depicting a wide variety of subjects, a randomly sampled batch may contain significantly diverse images. In that case, it becomes trivial for the model to distinguish between the positive and negative examples, thereby reducing the effectiveness of the learning process. To avoid this issue, we propose a strategy to guarantee the inclusion of hard negative examples in every batch, which is known to improve contrastive learning performance \cite{robinson2020contrastive, kalantidis2020hard}. Specifically, we first perform an a priori K-Means clustering \cite{macqueen1967some} of the visual features corresponding to the images comprising $\mathcal{D}$. Then, during training, we structure each batch such that a proportion $\alpha$ consists of images from the same cluster, ensuring the presence of hard negative examples. The remaining fraction, $(1 - \alpha)$, is filled with images randomly selected from the dataset. This approach strikes a balance by introducing challenging examples into the batch while also preserving a broad diversity within the images. This strategy differs from the one we used in the conference version of this work \cite{baldrati2023zero}, where we simply sampled the images within each batch at random. By including visually resembling examples within each batch, we encourage the model to focus on fine-grained details, improving its ability to discriminate between similar images. Consequently, as shown by the experimental results, the proposed hard negative sampling strategy improves the performance, especially on a dataset with a narrow domain such as FashionIQ \cite{wu2021fashion} (see \cref{sec:ablation_phi} for more details).

Differently from the conference version of this work \cite{baldrati2023zero}, we employ a combination of two losses to regularize the training of $\phi$. First, we employ the same $\mathcal{L}_{gpt}$ loss described in \cref{sec:optimization}. Second, inspired by \cite{gal2023encoder}, we propose to use an additional regularization penalty term to constrain the norm of the predicted pseudo-word tokens:
\begin{equation}
    \mathcal{L}_{pen} = \frac{1}{B} \sum_{i=1}^{B} \lVert v_{*}^{i} \rVert ^{2}_{2}
\end{equation}
This loss contributes to preventing the pseudo-word tokens generated by $\phi$ from residing in sparse regions of the CLIP token embedding space \cite{gal2023encoder}. From another point of view, $L_{pen}$ can be interpreted as a weight decay regularization term that is applied to the output of the network instead of its parameters. This aligns with the proposed OTI approach, where we apply weight decay regularization directly to the pseudo-word tokens. 

The final loss for training $\phi$ is
\begin{equation}\label{eq:loss_phi}    
    \mathcal{L}_{\phi} = \lambda_{distil} \mathcal{L}_{distil} +  \lambda_{\phi gpt} \mathcal{L}_{gpt} + \lambda_{pen} \mathcal{L}_{pen}
\end{equation}
with $\lambda_{distil}$, $\lambda_{\phi gpt}$, and $\lambda_{pen}$ representing the loss weights.

Since we do not leverage any labeled data, the training of our textual inversion network $\phi$ is entirely unsupervised. Indeed, differently from PALAVRA \cite{cohen2022this} and Context-I2W \cite{tang2023context}, we do not require any caption and employ only raw images. Specifically, we use the unlabeled test split of the ImageNet1K \cite{russakovsky2015imagenet} dataset as $\mathcal{D}$ to pre-train $\phi$. It comprises 100K images without any associated labels. Compared to Pic2Word \cite{saito2023pic2word} and Context-I2W \cite{tang2023context}, our method uses about 3\% of the data. We selected this dataset because it contains real-world images spanning a wide variety of subjects. The experiments show that our method is robust to the choice of the $\phi$ pre-training dataset (see \cref{sec:additional_experiments} for more details).

\section{CIRCO dataset} \label{sec:circo_dataset}
We recall that CIR datasets comprise triplets $(I_r, T_r, I_t)$ composed of a reference image, relative caption, and target image (\ie the ground truth), respectively. 

Existing datasets often include numerous false negatives, namely images that could potentially serve as valid ground truths for a query but are not labeled as such. This issue arises because, in each query triplet, only one image is designated as the target, rendering all other images as negatives. Additionally, most datasets are confined to specialized domains, such as fashion~\cite{berg2010automatic, han2017automatic, wu2021fashion, guo2018dialog}, birds~\cite{forbes2019neural}, or synthetic objects~\cite{vo2019composing}.
To the best of our knowledge, the CIRR dataset~\cite{liu2021image} is the sole dataset built on real-life images across an open domain. During the data collection process of CIRR, sets of 6 visually similar images are automatically generated. Subsequently, queries are devised so that both the reference and the target images belong to the same set, aiming to avoid the presence of false negatives within that particular set. However, this strategy does not guarantee the absence of false negatives throughout the entire dataset. Moreover, despite the visual similarity, the differences between images within the same set may not be easily expressible through relative captions and might necessitate absolute descriptions. This diminishes the significance of the visual information of the reference image and makes the retrieval task addressable with standard text-to-image techniques. For more details, refer to \cref{sec:quantitative_results}.

\begin{figure}
    \centering
    \includegraphics[width=\columnwidth]{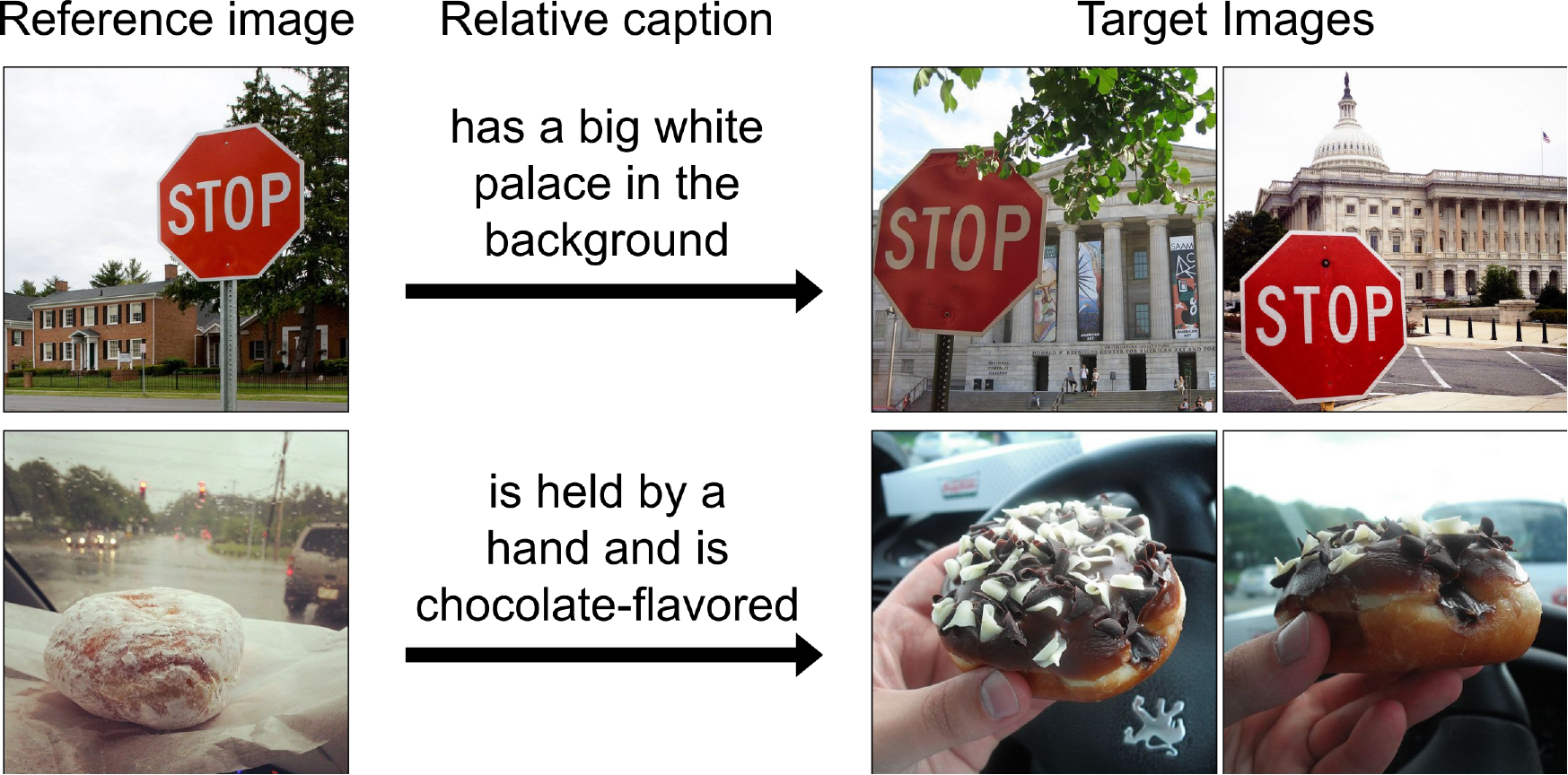}
    \caption{Examples of CIR queries and ground truths in CIRCO.}
    \label{fig:circo_example}
\end{figure}

To address these issues, we introduce an open-domain benchmarking dataset named CIRCO (Composed Image Retrieval on Common Objects in context). CIRCO is based on open-domain real-world images and is the first dataset for CIR with multiple ground truths and fine-grained semantic annotations. The whole annotation process has been carried out by the authors of this paper. To this end, we have developed a custom annotation tool that met our needs. The annotation process consists of three phases. In the first one, we build the triplets composed of a reference image, a relative caption, and a single target image. In the second one, we extend each triplet by annotating additional ground truths. In the third one, we assign semantic aspects to each query based on the relative caption. \Cref{fig:circo_example} shows some query examples of CIRCO.

\subsection{Triplets Annotation}\label{sec:triplets_annotation}
CIRCO is based on images sourced from COCO 2017 \cite{lin2014microsoft} unlabeled set, which comprises 123,403 images. This dataset was suitable for our goals as it comprises open-domain real-world images that portray a wide range of subjects. In addition, we opted for the COCO unlabeled set over the training one to avoid any pre-existing model biases, as the latter is commonly used for pre-training. In the COCO labeled sets, each object in an image is categorized under one of 12 supercategories: \textit{person}, \textit{animal}, \textit{sports}, \textit{vehicle}, \textit{food}, \textit{accessory}, \textit{electronic}, \textit{kitchen}, \textit{furniture}, \textit{indoor}, \textit{outdoor}, and \textit{appliance}.

The first step is leveraging CLIP ViT-L/14 zero-shot classification capabilities to associate every image of the unlabeled set to a supercategory. We assume that the classification is based on the predominant subject of each image.
This categorization aims to get an estimation of the content of each image to be able to later create a balanced dataset. Indeed, we annotate CIRCO so that the queries comprise reference images that are evenly distributed across the supercategories. 
This balancing step is required to address the noticeable domain bias observed in COCO images. Indeed, certain objects, such as stop signs and fire hydrants, are over-represented. 

The annotation tool selects a reference image at random and presents it alongside a gallery of 50 candidate target images. The target images must be visually similar to the reference one yet exhibit discernible disparities, as CIR requires the differences between them to be describable with a relative caption. Consequently, we choose the candidate target images based on their visual similarity to the reference image as per the CLIP features. To prevent the inclusion of near-identical images, we exclude those with a cosine similarity exceeding 0.92. The annotators are allowed to skip the current reference image if no suitable target is found in the gallery. On the contrary, when a suitable target image is available, the annotator selects it and writes the \textit{shared concept}, which represents the common characteristics between the reference and target images. We collect the shared concept to address any potential ambiguities. Finally, the annotator crafts a relative caption from the prefix ``Unlike the provided image, I want a photo of \{shared concept\} that". Since our goal is to create a challenging dataset comprising truly relative captions, we ensure that they are formulated in such a way that avoids references to subjects mentioned in the shared concept. In this way, the subject of the relative caption needs to be deduced from the reference image alone. 

At the end of this phase, we obtain 1020 triplets comprising a reference image, a relative caption, and a single target image.

\subsection{Multiple Ground Truths Annotation}\label{sec:multiple_gt_annotation}
For each triplet, we aim to label as ground truth all the images -- besides the target one -- that represent valid matches for the corresponding query. Given the starting triplet, the annotator needs to identify the ground truths from a gallery of images. 

We propose to facilitate the annotation process by exploiting our approach to retrieve the images from which the ground truths are selected. In particular, we employ \methodiccv to generate the pseudo-word $S_*$ corresponding to the reference image. Then, we carry out text-to-image retrieval based on the query `a photo of \{shared concept\} $S_*$ that \{relative caption\}". During the annotation phase, we incorporate the shared concept into the query because it improves performance. Indeed, considering the single ground truth triplets obtained in \cref{sec:triplets_annotation}, we achieve a Recall$@100$ of 82.15 with the shared concept and of 66.25 without it. In the gallery of images used for selecting the multiple ground truths, the annotation tool presents the top 100 retrieved images using our approach, along with the top 50 images most visually similar to the target one.

At the end of this phase, we have 4624 ground truths, of which 4097 were retrieved employing our method and 527 using the similarity with the target image. Since \methodiccv achieves a Recall$@100$ of 82.15, by approximation, we estimate that about 82.15\% of the total ground truths are present in the top 100 retrieved images. Consequently, the estimated total number of ground truths in our dataset is approximately $4097 / 0.8215 \approx 4,987$. Given that we labeled 4624 images as ground truth, we can infer that the annotated ones are $4,624 / 4,987 \approx 92.7\%$ of the total. Therefore, we estimate that our annotation strategy allows us to reduce the percentage of missing ground truths in the dataset to less than 10\%. 

Thanks to this second annotation step, we labeled additional $4624\!-\!1020\!=\!3604$ ground truths that would have otherwise been considered false negatives. Furthermore, this phase enables us to estimate the percentage of missing ground truths within the dataset. Notably, this estimation is unfeasible for CIR datasets featuring only a single ground truth, such as FashionIQ \cite{wu2021fashion} and CIRR \cite{liu2021image}, as they lack any information about the total number of ground truths. Indeed, in these datasets, the annotation process concludes upon the completion of the triplet construction.

 \begin{figure*}
  \centering
  \captionsetup[subfloat]{labelfont=normalfont, font=small, textfont=normalfont,skip=2pt} 
  \subfloat[Cardinality]{\includegraphics[width=0.3\linewidth]{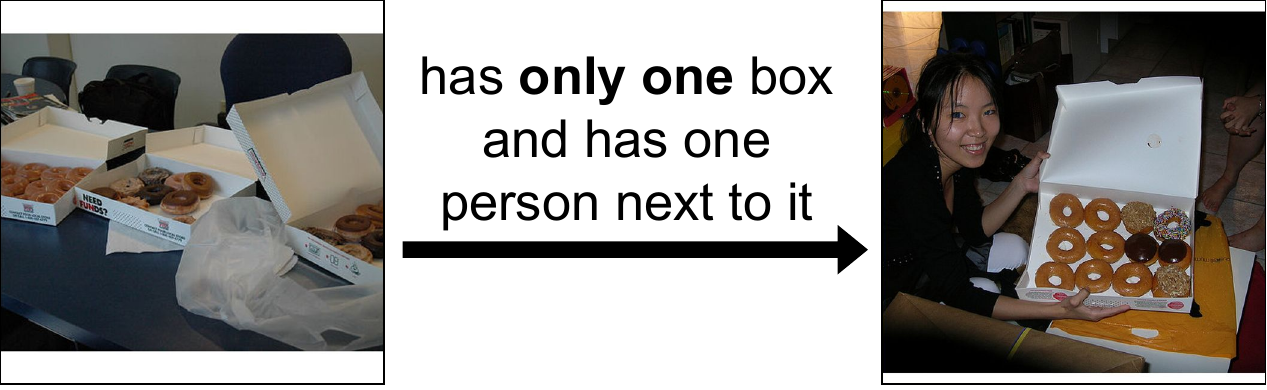}\label{fig:circo_example_cardinality}}
  \hfill
  \subfloat[Addition]{\includegraphics[width=0.3\linewidth]{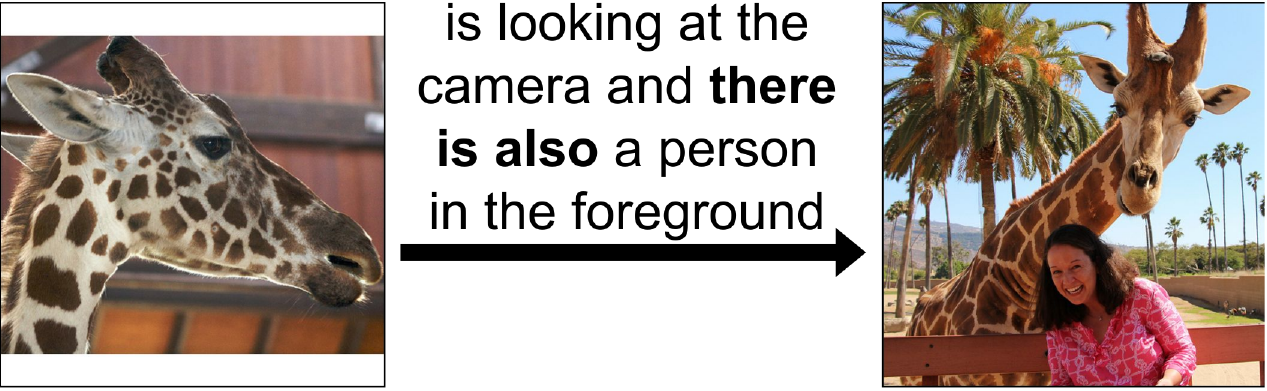}\label{fig:circo_example_addition}}
  \hfill
  \subfloat[Negation]{\includegraphics[width=0.3\linewidth]{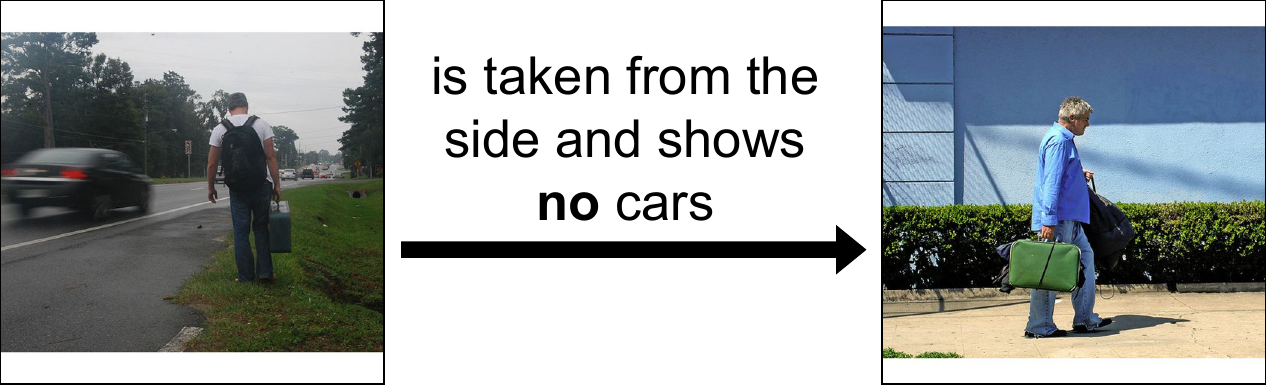}\label{fig:circo_example_negation}}
  
  \subfloat[Direct Addressing]{\includegraphics[width=0.3\linewidth]{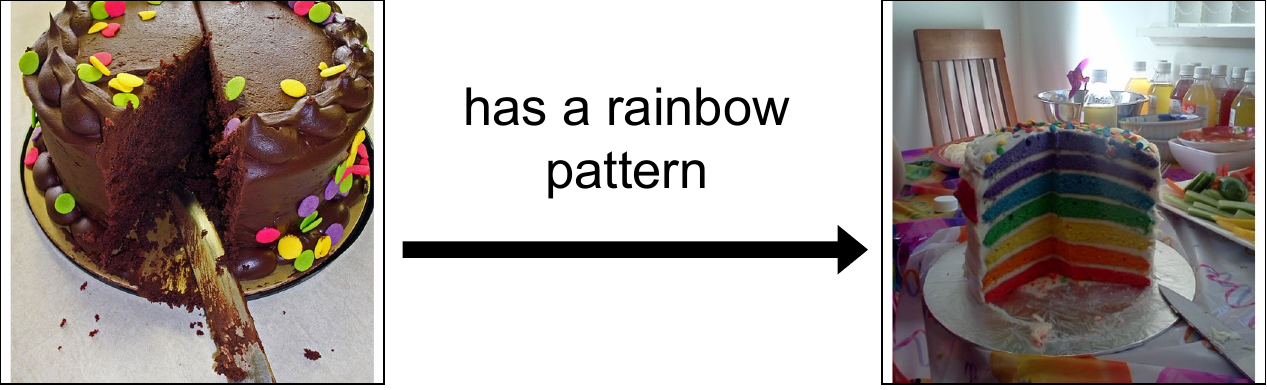}\label{fig:circo_example_direct_addressing}}
  \hfill
  \subfloat[Compare \& Change]{\includegraphics[width=0.3\linewidth]{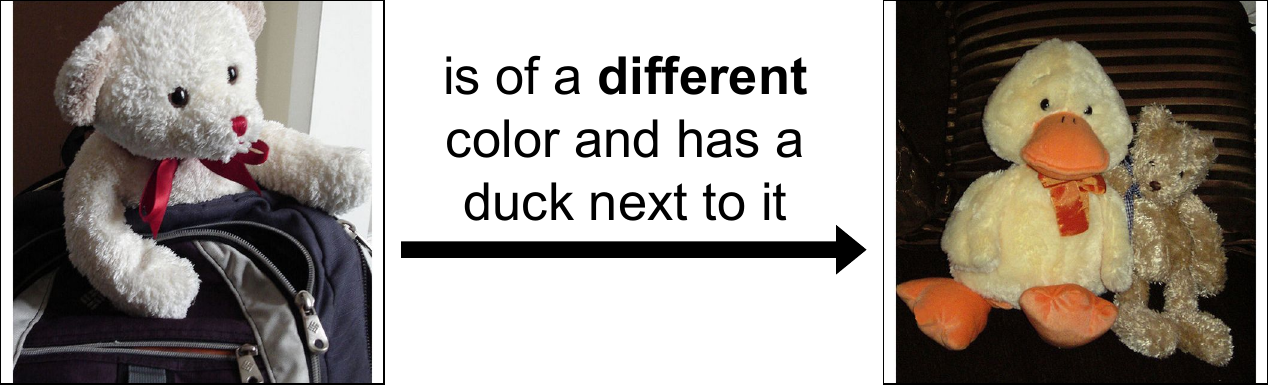}\label{fig:circo_example_compare_change}}
  \hfill
  \subfloat[Comparative Statement]{\includegraphics[width=0.3\linewidth]{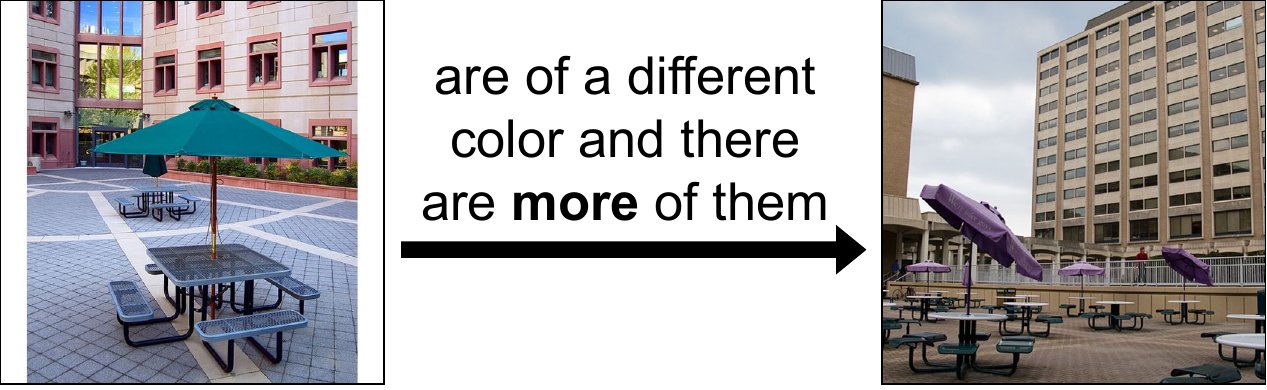}\label{fig:circo_example_comparative_statement}}
  
  \subfloat[Statement with Conjunction]{\includegraphics[width=0.3\linewidth]{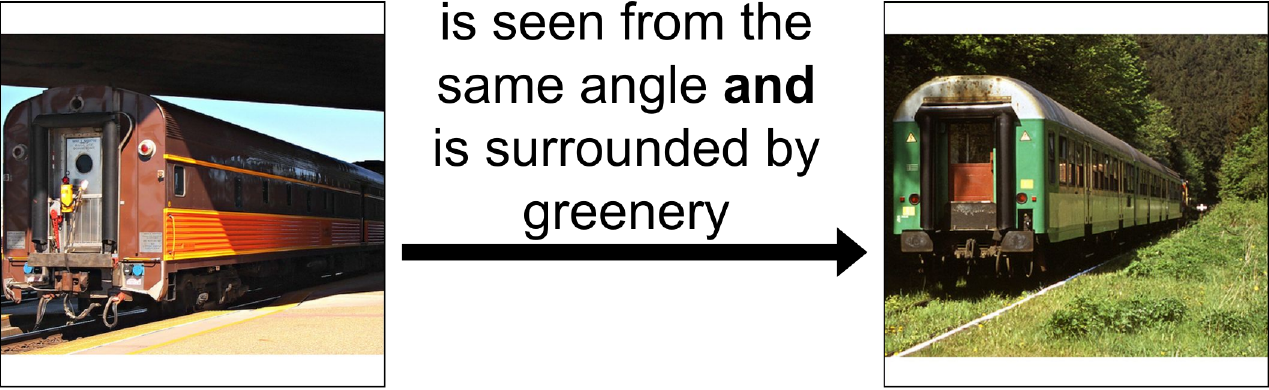}\label{fig:circo_example_statement_conjunction}}
  \hfill
  \subfloat[Spatial Relations \& Background]{\includegraphics[width=0.3\linewidth]{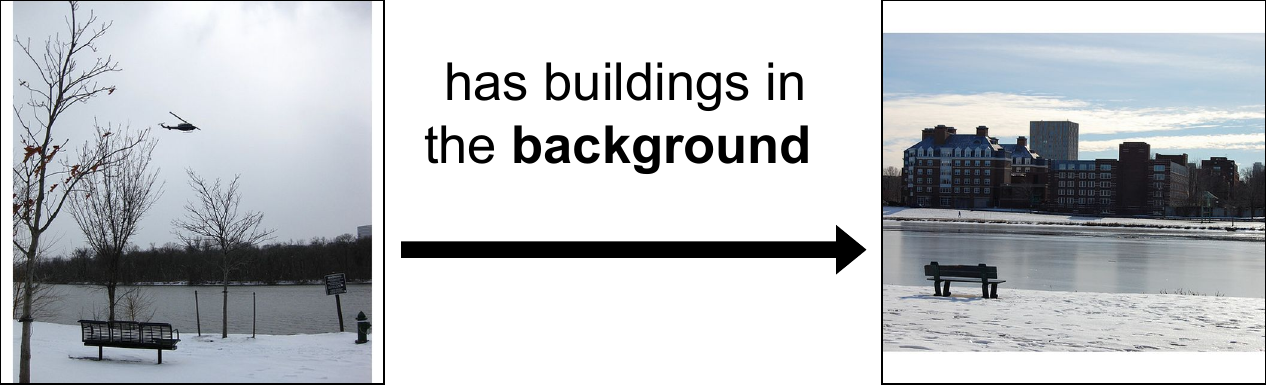}\label{fig:circo_example_background}}
  \hfill
  \subfloat[Viewpoint]{\includegraphics[width=0.3\linewidth]{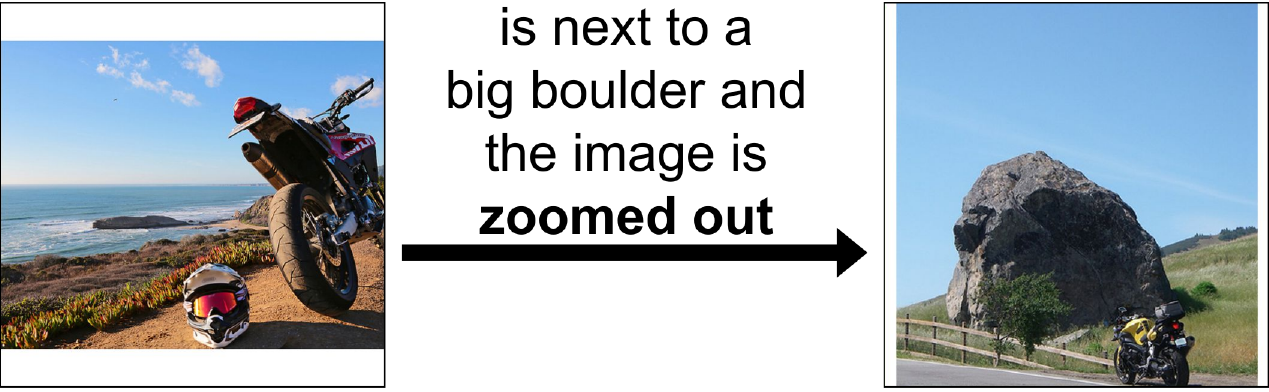}\label{fig:circo_example_viewpoint}}
  \caption{Examples of queries of the proposed CIRCO dataset for different semantic aspects. For simplicity, we report only one ground truth. We highlight the keywords of each semantic aspect in bold.}
  \label{fig:circo_semantic_examples}
\end{figure*}

\subsection{Semantic Aspects Annotation}

To allow a fine-grained semantic analysis on CIRCO, we perform a third annotation phase to assign semantic aspects to each query depending solely on its relative caption. In particular, we consider the same semantic categories as CIRR \cite{liu2021image}, such as ``\textit{direct addressing}" and ``\textit{compare \& change}".

A preliminary labeling of the semantic aspects was carried out during the multiple ground truths annotation phase in the conference version of this work \cite{baldrati2023zero}. The goal was to measure some raw statistics on the semantic categories of the relative captions of CIRCO. In that case, the semantic aspects were determined only by the user responsible for annotating the corresponding query. However, given the broad terms used to refer to the categories and the ambiguity of language, some inconsistencies in the classification among the annotators are inevitable. Given that we no longer aim just to compute raw statistics but rather conduct a fine-grained semantic analysis of the results on CIRCO, we require clean and reliable annotations. To this end, we introduce an additional annotation phase entirely focused on the semantic aspects related to the queries. First, we set a collection of rules on how to label each semantic category to remove possible ambiguities. In particular, a semantic aspect is assigned to a query if the relative caption: a) \textit{cardinality}: explicitly mentions a specific number of objects in the scene, \eg one, two; b) \textit{addition}: requires the addition of an object that is not present in the reference image, \eg shows also a cat; c) \textit{negation}: requests the removal of an object present in the reference image, \eg is without cars; d) \textit{direct addressing}: explicitly requests for a specific change, \eg is playing with a red ball; e) \textit{compare \& change}: requests to change something while mentioning an attribute of the reference image, \eg there is a cat instead of a dog; f) \textit{comparative statement}: includes a comparison, \eg has more people; g) \textit{statement with conjunction}: includes a conjunction proposition, \eg and, or; h) \textit{spatial relations \& background}: references the background or spatial relations among objects, \eg has a lake in the background; i) \textit{viewpoint}: mentions a specific viewpoint or perspective, \eg is shot from the top. Note that these categories are not mutually exclusive, \ie a single relative caption can be labeled with multiple semantic aspects. We provide a query example for each semantic aspect in \cref{fig:circo_semantic_examples}. Then, we make all the annotators label the semantic aspects of all the queries following the set of rules. Finally, we obtain the ground truth annotations by assigning a semantic aspect to a query if at least half of the annotators agree on the corresponding annotation. Therefore, differently from the conference version of this work \cite{baldrati2023zero}, each semantic category label stems from the judgment of multiple annotators and thus has a higher reliability.

After this third annotation phase, we obtain a clean and reliable semantic categorization of the queries. As a result, CIRCO is the first CIR dataset that enables a fine-grained semantic analysis of the performance of different methods. Indeed, existing datasets \cite{wu2021fashion, liu2021image} just report raw statistics of the semantic categorization of the queries. Moreover, such categorization is not publicly available, making a fine-grained semantic analysis of the performance unfeasible. In contrast, we believe that performing such an analysis of the results is crucial, as it allows us to identify the most challenging query types based on their semantic categories and thus foster focused research efforts.

\subsection{Dataset Analysis} \label{sec:dataset_analysis}
CIRCO comprises 1020 queries, randomly divided into 220 and 800 for the validation and test set, respectively. The total number of ground truths is 4624, i.e. 4.53 per query on average. 
The maximum number of ground truths annotated for a query is 21, while the modal value is 2. 

 \begin{table}
     \caption{Analysis of the semantic aspects covered by the relative captions. $^{\dagger}$ indicates results taken from \cite{liu2021image}. -- denotes no reported results.}
     \centering
     \Large
     \resizebox{\linewidth}{!}{ 
     \begin{tabular}{lccc} 
       \toprule
       \multicolumn{1}{l}{\multirow{2}{*}{Semantic Aspect}} & \multicolumn{3}{c}{Coverage (\%)} \\ 
       & \multicolumn{1}{c}{CIRCO} & \multicolumn{1}{c}{CIRR} & \multicolumn{1}{c}{FashionIQ} \\
       \midrule
       Cardinality & 16.3 & 29.3$^\dagger$ & -- \\
       Addition & 36.6 & 15.2$^\dagger$ & 15.7$^\dagger$ \\
       Negation & 11.0 & 11.9$^\dagger$ & 4.0$^\dagger$ \\ 
       Direct Addressing & 54.2 & 57.4$^\dagger$ & 49.0$^\dagger$ \\
       Compare \& Change & 37.8 & 31.7$^\dagger$ & ~3.0$^\dagger$ \\
       Comparative Statement & 25.7 & 51.7$^\dagger$ & 32.0$^\dagger$ \\
       Statement with Conjunction & 76.2 & 43.7$^\dagger$ & 19.0$^\dagger$ \\ 
       Spatial Relations \& Background & 46.5 & 61.4$^\dagger$ & -- \\
       Viewpoint & 22.1 & 12.7$^\dagger$ & -- \\
       \midrule
       \textit{Avg. Caption Length (words)} & 10.4 & 11.3$^\dagger$ & 5.3$^\dagger$ \\
       \bottomrule
       \end{tabular}}
   \label{tab:caption_semantic}
 \end{table}

\begin{figure}
     \centering
     \includegraphics[width=\linewidth]{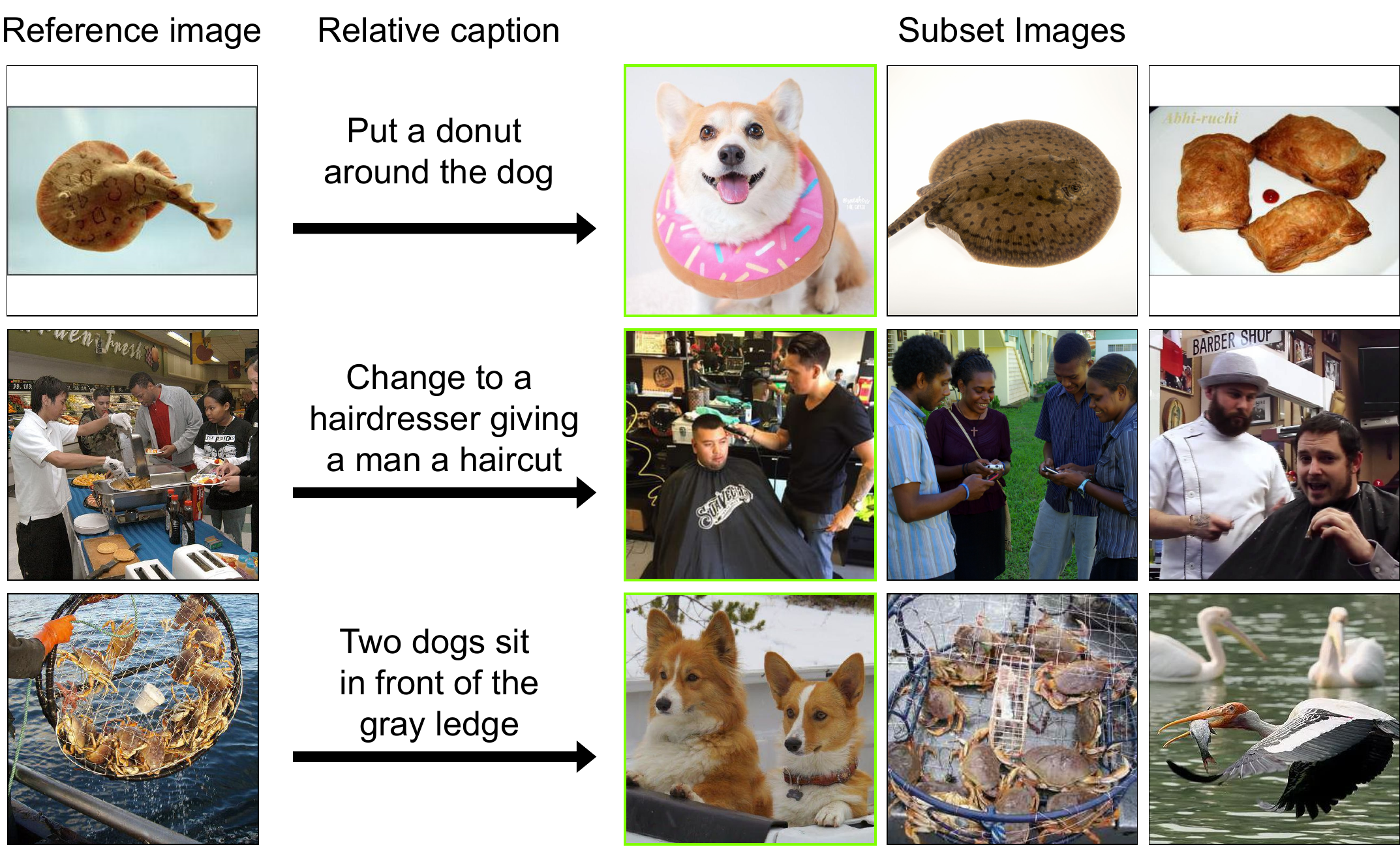}
     \caption{Examples of queries belonging to the CIRR dataset \cite{liu2021image}. The subset images depict very different subjects and the relative captions do not consider the reference images. We highlight the target image with a green border.}
     \label{fig:cirr_subset_example}
 \end{figure}

The relative captions consist of an average of 10.4 words. Similar to CIRR \cite{liu2021image}, in \cref{tab:caption_semantic} we report the raw statistics related to the semantic aspects associated with the relative captions. We observe that the average length of the captions and overall coverage of the semantic categories are comparable with CIRR. However, in CIRCO approximately 75\% of the annotations are composed of multiple statements, more than the $\sim$43\% of CIRR, thus revealing a higher complexity.

\begin{figure}
 \centering
     \includegraphics[width=0.85\linewidth]{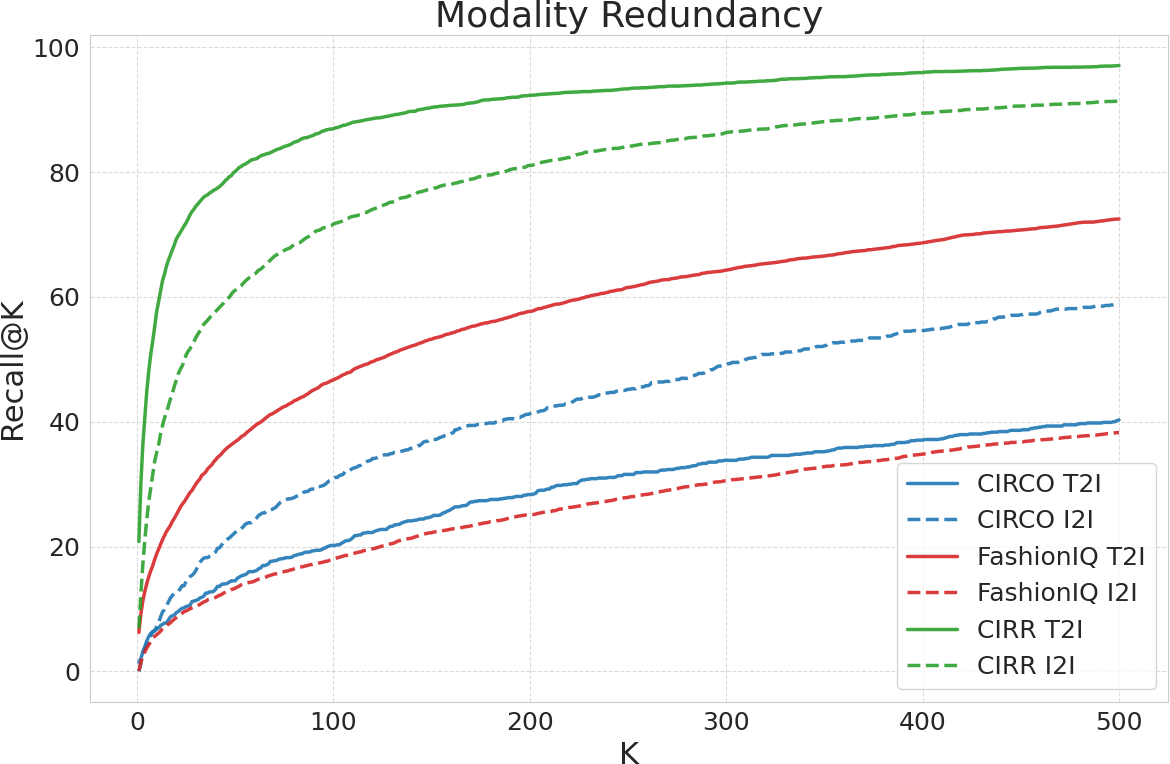}
 \caption{Evaluation of the modality redundancy of CIRCO, CIRR, and FashionIQ validation sets. Lower values are better. T2I and I2I represent Text-to-Image and Image-to-Text retrieval, respectively.}
 \label{fig:modality_redundancy}
\end{figure}

CIRR \cite{liu2021image} validation and test sets comprise 4K triplets each. We remind that during the data collection process, CIRR automatically assembles subsets of 6 visually similar images based on the features of a ResNet152 \cite{he2016deep}. Then, the queries are formulated to ensure that the reference and the target images belong to the same subset. However, despite the feature similarity, the images within these subsets often portray significantly different subjects. As a result, it becomes impossible for a human annotator to craft a relative caption, and they need to resort to an absolute description of the target image. \Cref{fig:cirr_subset_example} shows some examples of this problem. We observe that, for instance, the annotator needs to rely on an absolute caption to describe the differences between an image depicting a crab fisherman and one with two dogs. To address this issue, we design an annotation strategy for CIRCO that lets the annotators choose the reference-target pair without constraints. As a result, we ensure that the annotators only craft captions that are truly relative, thus enhancing the quality of the dataset. To confirm this, similar to \cite{levy2023data}, we carry out an experiment to evaluate whether both the reference image and the relative caption are necessary for retrieval. Specifically, we quantitatively assess the degree of redundancy of each of the two modalities (\ie image and text) by measuring the Recall@K performance of Text-to-Image (T2I) and Image-to-Image (I2I) retrieval, using respectively the relative caption and the reference image as the query. Indeed, high T2I performance implies that the relative captions are actually absolute and that the reference images are redundant. On the contrary, strong I2I results mean that the reference and target images are very similar, making the relative caption redundant. \Cref{fig:modality_redundancy} shows the results for varying K values. For a fair comparison with single ground truth datasets such as CIRR and FashionIQ, for CIRCO we consider only the single ground truth annotated during the first phase (\cref{sec:triplets_annotation}). A lower curve suggests that the corresponding dataset is more difficult for a uni-modal query thus indicating a lower modality redundancy. Compared to CIRR, CIRCO demonstrates significantly lower recall metrics for both T2I and I2I, proving the quality of the proposed annotation strategy. In addition, CIRCO obtains comparable results to FashionIQ, despite encompassing a considerably broader domain.

Compared to CIRR, CIRCO comprises fewer queries, but our three-phase annotation strategy ensures higher quality, reduced false negatives, the availability of multiple ground truths, and public and reliable semantic annotations. Moreover, since its introduction in the conference version of this work \cite{baldrati2023zero}, CIRCO has been recognized as the CIR dataset with the highest quality \cite{li2024catllm} and the cleanest annotations \cite{karthik2024visionbylanguage}. Finally, we employ all the 120K images of COCO as the index set, thereby providing considerably more distractors than the 2K images of the CIRR test set.

\begin{table*}[!ht]
  \centering
  \caption{Quantitative results on FashionIQ validation set. Best and second-best scores are highlighted in bold and underlined, respectively. $^{\dagger}$ indicates results from the original paper.}
  \resizebox{0.8\linewidth}{!}{ 
  \begin{tabular}{clcccccccc} 
  \toprule
  \multicolumn{1}{c}{} & \multicolumn{1}{c}{} & \multicolumn{2}{c}{Shirt} & \multicolumn{2}{c}{Dress} &\multicolumn{2}{c}{Toptee} &\multicolumn{2}{c}{Average} \\
  \cmidrule(lr){3-4}
  \cmidrule(lr){5-6}
  \cmidrule(lr){7-8}
  \cmidrule(lr){9-10}
  \multicolumn{1}{c}{Backbone} & \multicolumn{1}{l}{Method} & R$@10$ & R$@50$ & R$@10$ & R$@50$ & R$@10$ & R$@50$ & R$@10$ & R$@50$ \\ 
  \midrule
  \multirow{9}{*}{B/32} & Image-only & 6.92 & 14.23 & 4.46 & 12.19 & 6.32 & 13.77 & 5.90 & 13.37 \\
  & Text-only & 19.87 & 34.99 & 15.42 & 35.05 & 20.81 & 40.49 & 18.70 & 36.84 \\ 
  & Image + Text & 13.44 & 26.25 & 13.83 & 30.88 & 17.08 & 31.67 & 14.78 & 29.60 \\
  & Captioning & 17.47 & 30.96 & 9.02 & 23.65 & 15.45 & 31.26 & 13.98 & 28.62 \\
  & PALAVRA \cite{cohen2022this} & 21.49 & 37.05 & 17.25 & 35.94 & 20.55 & 38.76 & 19.76 & 37.25 \\
  & \methodiccv-OTI$^{\dagger}$ \cite{baldrati2023zero} & 25.37 & 41.32 & 17.85 & 39.91 & 24.12 & 45.79 & 22.44 & 42.34 \\
  & \methodiccv$^{\hspace{-2pt}\dagger}$ \cite{baldrati2023zero} & 24.44 & 41.61 & 18.54 & 39.51 & 25.70 & 46.46 & 22.89 & 42.53  \\ 
  \rowcolor{LightCyan} \cellcolor{white} & \textbf{\methodpami-OTI} & \textbf{27.09} & \underline{43.42} & \textbf{21.27} & \textbf{42.19} & \textbf{26.82} & \textbf{48.75} & \textbf{25.06} & \underline{44.79} \\
  \rowcolor{LightCyan} \cellcolor{white} & \textbf{\methodpami} & \underline{25.81} & \textbf{43.52} & \underline{20.92} & \textbf{42.19} & \underline{26.47} & \underline{48.70} & \underline{24.40} & \textbf{44.80} \\ 					
  \midrule[.02em]
  \multirow{7}{*}{L/14} & Pic2Word$^{\dagger}$ \cite{saito2023pic2word} & 26.20 & 43.60 & 20.00 & 40.20 & 27.90 & 47.40 & 24.70 & 43.70 \\
  & Context-I2W$^{\dagger}$ \cite{tang2023context} & 29.70 & \underline{48.60} & \underline{23.10} & \underline{45.30} & 30.60 & \underline{52.90} & \underline{27.80} & 48.93 \\
  & LinCIR$^{\dagger}$ \cite{gu2023language} & 29.10 & 46.81 & 20.92 & 42.44 & 28.81 & 50.18 & 26.28 & 46.49 \\
  & \methodiccv-XL-OTI$^{\dagger}$ \cite{baldrati2023zero} & \underline{30.37} & 47.49 & 21.57 & 44.47 & 30.90 & 51.76 & 27.61 & 47.90 \\
  & \methodiccv-XL$^{\dagger}$ \cite{baldrati2023zero} & 26.89 & 45.58 & 20.48 & 43.13 & 29.32 & 49.97 & 25.56 & 46.23 \\ 
  \rowcolor{LightCyan} \cellcolor{white}& \textbf{\methodpami-XL-OTI} & \textbf{31.80} & \textbf{50.20} & \textbf{24.19} & 45.12 & \textbf{31.72} & \textbf{53.29} & \textbf{29.24} & \textbf{49.54} \\
  \rowcolor{LightCyan} \cellcolor{white} & \textbf{\methodpami-XL} & 28.75 & 47.84 & 22.51 & \textbf{46.36} & \underline{31.31} & 52.68 & 27.52 & \underline{48.96} \\ 
  \bottomrule
  \end{tabular}}
  \label{tab:fashioniq_val}
\end{table*}

\subsection{Evaluation Metric}
To alleviate the problem of false negatives, most works evaluate the performance on CIR datasets using Recall@K, with K set to quite large values (\eg 10, 50 \cite{wu2021fashion}). This makes a fine-grained analysis of the models difficult.

Thanks to the reduced false negatives and multiple ground truths of CIRCO, we can rely on a more fine-grained metric for performance evaluation, such as mean Average Precision (mAP). Indeed, mAP considers also the ranks in which the ground truths are retrieved. Specifically, we compute mAP$@K$, with K ranging from 5 to 50, as follows:
\begin{equation}
    \text{mAP}@K = \frac{1}{N} \sum\limits^N_{n=1} \frac{1}{\min(K, G_n)} \sum\limits_{k=1}^K P@k * \text{rel}@k
\end{equation}
where $N$ is the number of queries, $G_n$ is the number of ground truths of the $n$-th query, $P@k$ is the precision at rank $k$, rel$@k$ is a relevance function. The relevance function is an indicator function that equals 1 if the image at rank $k$ is labeled as a ground truth and equals 0 otherwise.

\section{Experimental Results} \label{sec:experimental_results}
We measure the performance of our method following the standard evaluation protocol \cite{baldrati2022conditioned, liu2021image} on the three main CIR datasets: FashionIQ \cite{wu2021fashion}, CIRR \cite{liu2021image} and the proposed CIRCO \cite{baldrati2023zero}. Specifically, we use the three categories of FashionIQ validation split and the test sets of CIRR and CIRCO. Moreover, we evaluate the performance of \methodpami on two additional settings, introduced in \cite{saito2023pic2word}: object composition on COCO \cite{lin2014microsoft} and domain conversion on ImageNet \cite{russakovsky2015imagenet, hendrycks2021many}. In this case, we follow the evaluation protocol adopted by \cite{saito2023pic2word, tang2023context}.

We present two variants of our method: \methodpami, based on CLIP ViT-B/32, and \methodpami-XL, using CLIP ViT-L/14 as the backbone. From now on, we will refer to ViT-B/32 and ViT-L/14 as B/32 and L/14, respectively.

\subsection{Implementation Details} \label{sec:implementation_details}
Regarding the Optimization-based Textual Inversion (OTI), we perform 500 iterations with a learning rate of $2e\!-\!2$. We set the loss weights $\lambda_{content}$ and $\lambda_{OTIgpt}$ in \cref{eq:loss_oti} to 1 and 0.5, respectively. We set the standard deviation $\gamma$ of the Gaussian noise to 0.64 and 0.16 respectively for \methodpami-OTI and \methodpami-XL-OTI. For the textual inversion network $\phi$, we train for 115 epochs, with a learning rate of $1e\!-\!4$ and a batch size of 256. We set the loss weights $\lambda_{distil}$ and $\lambda_{\phi gpt}$ in \cref{eq:loss_phi} to 1 and 0.75, respectively. The loss weight $\lambda_{pen}$ is equal to $3e\!-\!3$ and $1e\!-\!2$ for the B/32 and L/14 backbones, respectively. The temperature $\tau$ in \cref{eq:loss_distil} is set to 0.25. The parameter $\alpha$ of the hard negative sampling strategy (\cref{sec:phi_training}) is set to 0.5. For both OTI and $\phi$, we employ the AdamW optimizer \cite{loshchilov2018decoupled} with weight decay 0.01. We use an exponential moving average of 0.99 and 0.999 decay for OTI and $\phi$, respectively. During OTI, we set the number of concept words $k$ associated with each image to 15, while during the training of $\phi$ to 150. We tune each hyperparameter individually with a grid search on the CIRR validation set.

Using a single NVIDIA A100 40GB GPU, \methodpami-XL-OTI takes $\sim$35 seconds for a single image and $\sim$1.1 seconds per image with batch size 256. The training of $\phi$ for \methodpami-XL takes 12 hours in total on a single A100 GPU. Throughout all the experiments, we adopt the pre-processing technique introduced in \cite{baldrati2023composed}. For retrieval, we normalize both the query and index set features to have a unit $L_2$-norm.

To generate the phrases used for the regularization with $\mathcal{L}_{gpt}$, we exploit the GPT-Neo-2.7B model with 2.7 billion parameters developed by EleutherAI. For each of the 20,932 class names of the Open Images V7 dataset \cite{kuznetsova2020open}, we generate 256 phrases a priori with a temperature of 0.5 and a maximum length constraint of 35 tokens. The whole process requires about 12 hours to execute on a single A100 GPU. Since this operation only needs to be performed once, the time requirements are manageable.

We use the same set of templates during both training and inference. Since the FashionIQ dataset provides two relative captions per triplet, at inference time, we concatenate them using the conjunction ``and". To ensure our method remains invariant to the concatenation order, we use both possible concatenation orders and average the resulting features.

\subsection{Quantitative Results}\label{sec:quantitative_results}

We provide the results of both \methodpami and \methodpami-OTI. We compare our method with several zero-shot baselines: 1) \textit{Text-only}: we compute the similarity using only the CLIP features of the relative caption; 2) \textit{Image-only}: we retrieve the most similar images to the reference one; 3) \textit{Image + Text}: we sum together the CLIP features of the reference image and the relative caption; \textit{4) Captioning}: we substitute the pseudo-word token with the caption of the reference image obtained via a pre-trained captioning model~\cite{yu2022coca}\footnote{\href{https://huggingface.co/laion/CoCa-ViT-B-32-laion2B-s13B-b90k}{https://huggingface.co/laion/CoCa-ViT-B-32-laion2B-s13B-b90k}}. In addition, we compare the proposed approach with the previous version of our method \cite{baldrati2023zero} and state-of-the-art ZS-CIR methods: Pic2Word \cite{saito2023pic2word}, Context-I2W \cite{tang2023context}, and LinCIR \cite{gu2023language}. For a fair comparison, we only consider competing methods that rely on the CLIP model without fine-tuning its weights and that, at inference time, do not require any additional pre-trained models, such as LLMs. 

\begin{table}[!t]
  \centering
  \Large
  \caption{Quantitative results on CIRR test set. $^{\dagger}$ indicates results from the original paper. -- denotes results not reported in the original paper.}
  \resizebox{\linewidth}{!}{ 
  \begin{tabular}{clcccc} 
  \toprule
  \multicolumn{1}{c}{} & \multicolumn{1}{c}{} & \multicolumn{4}{c}{Recall$@K$} \\
  \cmidrule(lr){3-6}
  \multicolumn{1}{l}{Backbone} & \multicolumn{1}{l}{Method} & $K = 1$ & $K = 5$ & $K = 10$ & $K = 50$ \\ 
  \midrule
  \multirow{9}{*}{B/32} & Image-only & 6.89  & 22.99  & 33.68 & 59.23 \\ 
  & Text-only & 21.81 & 45.22 & 57.42 & 81.01 \\ 
  & Image + Text & 11.71 & 35.06 & 48.94 & 77.49 \\
  & Captioning & 12.46 & 35.04 & 47.71 & 77.35 \\
  & PALAVRA \cite{cohen2022this} & 16.62 & 43.49 & 58.51 & 83.95 \\
  & \methodiccv-OTI$^{\dagger}$ \cite{baldrati2023zero} & 24.27 & 53.25 & 66.10 & 88.84 \\
  & \methodiccv$^{\hspace{-2pt}\dagger}$ \cite{baldrati2023zero} & 24.00  & 53.42 & 66.82 & 89.78 \\
  \rowcolor{LightCyan} \cellcolor{white} & \textbf{\methodpami-OTI} & \textbf{26.19} & \underline{55.18} & \textbf{68.55} & 90.65 \\
  \rowcolor{LightCyan} \cellcolor{white} & \textbf{\methodpami} & \underline{25.23} & \textbf{55.69} & \underline{68.05} & \textbf{90.82} \\ \midrule[.02em]
  \multirow{7}{*}{L/14} & Pic2Word$^{\dagger}$ \cite{saito2023pic2word} & 23.90 & 51.70 & 65.30 & 87.80 \\
  & Context-I2W$^{\dagger}$ \cite{tang2023context} & \textbf{25.60} & \textbf{55.10} & \textbf{68.50} & \textbf{89.80} \\
  & LinCIR$^{\dagger}$ \cite{gu2023language} & 25.04 & 53.25 & 66.68 & -- \\
  & \methodiccv-XL-OTI$^{\dagger}$ \cite{baldrati2023zero} & 24.87 & 52.31 & 66.29 & 88.58 \\
  & \methodiccv-XL$^{\dagger}$ \cite{baldrati2023zero} & 24.24 & 52.48 & 66.29 & 88.84 \\
  \rowcolor{LightCyan} \cellcolor{white} & \textbf{\methodpami-XL-OTI} & \underline{25.40} & \underline{54.05} & \underline{67.47} & \underline{88.92} \\
  \rowcolor{LightCyan} \cellcolor{white} & \textbf{\methodpami-XL} & 25.28 & 54.00 & 66.72 & 88.80 \\
  \bottomrule
  \end{tabular}}
  \label{tab:cirr_test}
\end{table}

\tit{FashionIQ}
We report the results for FashionIQ in \cref{tab:fashioniq_val}. Considering the B/32 backbone, \methodpami obtains comparable performance to \methodpami-OTI, thereby preserving effectiveness while offering a notable efficiency improvement. Both versions of our approach outperform the baselines, including the preliminary version of this work \cite{baldrati2023zero}. Notably, the improvement over Captioning highlights that the pseudo-word token encapsulates more information than the actual words forming the generated caption. Regarding the L/14 backbone, we notice that, despite using only 3\% of the training data, \methodpami-XL achieves a considerable performance improvement over Pic2Word and comparable results with Context-I2W.
In addition, we recall that Context-I2W requires a double forward pass of the text encoder and employs a transformer-based architecture significantly more complex than our MLP-based one. We observe a performance gap compared to \methodpami-XL-OTI. We suppose that this discrepancy may stem from the very narrow domain of FashionIQ, which differs considerably from the natural images of the pre-training dataset we employ for training $\phi$. To support this hypothesis, we trained a version of \methodpami-XL using the FashionIQ training set as the pre-training dataset, yielding an average Recall@10 and Recall@50 of 29.07 and 49.67, respectively, on the validation set. These results closely align with those of \methodpami-XL-OTI, confirming our theory. We provide more details on the impact of the $\phi$ pre-training dataset in \cref{sec:additional_experiments}.

\begin{table}[]
  \centering
  \caption{Quantitative results on CIRCO test set. $^{\dagger}$ indicates results from the original paper.}
  \Large
  \resizebox{\linewidth}{!}{ 
  \begin{tabular}{clcccc} 
  \toprule
  \multicolumn{1}{c}{} & \multicolumn{1}{c}{} & \multicolumn{4}{c}{mAP$@K$} \\
  \cmidrule(lr){3-6}
  \multicolumn{1}{l}{Backbone} & \multicolumn{1}{l}{Method} & $K=5$ & $K=10$ & $K=25$ & $K=50$ \\ \midrule
  \multirow{9}{*}{B/32} & Image-only & 1.34 & 1.60 & 2.12 & 2.41 \\
  & Text-only & 2.56 & 2.67 & 2.98 & 3.18 \\
  & Image + Text & 2.65 & 3.25 & 4.14 & 4.54 \\
  & Captioning & 5.48 & 5.77 & 6.44 & 6.85 \\
  & PALAVRA \cite{cohen2022this} & 4.61 & 5.32 & 6.33 & 6.80 \\
  & \methodiccv-OTI$^{\dagger}$ \cite{baldrati2023zero} & 7.14 & 7.83 & 8.99 & 9.60 \\
  & \methodiccv$^{\hspace{-2pt}\dagger}$ \cite{baldrati2023zero} & 9.35 & 9.94 & 11.13 & 11.84 \\
  \rowcolor{LightCyan} \cellcolor{white} & \textbf{\methodpami-OTI} & \underline{10.31} & \underline{10.94} & \underline{12.27} & \underline{13.01} \\
  \rowcolor{LightCyan} \cellcolor{white} & \textbf{\methodpami} & \textbf{10.58} & \textbf{11.24} & \textbf{12.51} & \textbf{13.26} \\ \midrule[.02em]
  \multirow{6}{*}{L/14} & Pic2Word~\cite{saito2023pic2word} & 8.72 & 9.51 & 10.64&11.29 \\
  & LinCIR$^{\dagger}$ \cite{gu2023language} & \textbf{12.59} & \underline{13.58} & \underline{15.00} & \underline{15.85} \\
  & \methodiccv-XL-OTI$^{\dagger}$ \cite{baldrati2023zero} & 10.18 & 11.03 & 12.72 & 13.67 \\
  & \methodiccv-XL$^{\dagger}$ \cite{baldrati2023zero} & 11.68  & 12.73 & 14.33 & 15.12 \\
  \rowcolor{LightCyan} \cellcolor{white} & \textbf{\methodpami-XL-OTI} & 11.31 & 12.67 & 14.46 & 15.34 \\
  \rowcolor{LightCyan} \cellcolor{white} & \textbf{\methodpami-XL} & \underline{12.50} & \textbf{13.61} & \textbf{15.36} & \textbf{16.25} \\
  \bottomrule
  \end{tabular}}
  \label{tab:circo_test}
\end{table}

\tit{CIRR}
\Cref{tab:cirr_test} shows the results for the CIRR test set. We notice that the Text-only baseline outperforms Image-only and Image+Text. These results reveal a major issue with CIRR: the relative captions are often not truly relative in practice. In particular, as observed also in \cite{saito2023pic2word}, we notice that the reference image may not provide useful information for retrieval and may even have a detrimental effect. 

We observe that, for both backbones, the results achieved by our method with OTI and $\phi$ are comparable, thereby proving the effectiveness of the proposed distillation process. Interestingly, there is no performance gap between the B/32 and L/14 versions, and actually, the B/32 even outperforms the L/14 in most cases. When compared with the conference version of this work \cite{baldrati2023zero}, our method obtains better results for both the OTI and $\phi$ versions, highlighting the importance of the improvements introduced in this work. Regarding the L/14 backbone, Context-I2W \cite{tang2023context} achieves the best performance. However, such a method is trained on the CC3M \cite{sharma2018conceptual} dataset, which is more than 30 times larger than our pre-training dataset. When considering half of the training images, the authors of Context-I2W report significantly lower results, with a $R@1$ and $R@5$ of 24.80 and 53.60, respectively. Therefore, despite using 15 times fewer data and no captions, \methodpami-XL still obtains comparable performance, with a $R@1$ and $R@5$ of 25.28 and 54.00, respectively.

\begin{table*}[!t]
  \centering
  \caption{Quantitative results on CIRCO test set for each semantic category.} 
  \Large
  \resizebox{\linewidth}{!}{ 
  \begin{tabular}{lcc>{\columncolor{LightCyan}}c>{\columncolor{LightCyan}}c|ccc>{\columncolor{LightCyan}}c>{\columncolor{LightCyan}}c} 
  \toprule
  \multicolumn{1}{c}{} & \multicolumn{4}{c}{ViT-B/32} & \multicolumn{5}{c}{ViT-L/14} \\
  \cmidrule(lr){2-5}
  \cmidrule(lr){6-10}
  \multicolumn{1}{l}{Semantic Aspect} & \multicolumn{1}{c}{PALAVRA} & \multicolumn{1}{c}{\methodiccv} & \multicolumn{1}{c}{\cellcolor{LightCyan}\textbf{\methodpami-OTI}} & \multicolumn{1}{c}{\cellcolor{LightCyan}\textbf{\methodpami}} & \multicolumn{1}{c}{Pic2Word} & \multicolumn{1}{c}{LinCIR} & \multicolumn{1}{c}{\methodiccv-XL} & \multicolumn{1}{c}{\cellcolor{LightCyan}\textbf{\methodpami-XL-OTI}} & \multicolumn{1}{c}{\cellcolor{LightCyan}\textbf{\methodpami-XL}} \\ \midrule
Cardinality                    & 3.38 & 7.94  & \underline{8.29}  & \textbf{9.50}   & 9.20 & \underline{11.80}  & 10.25 & 10.30  & \textbf{11.59} \\
Addition                       & 5.66 & 10.55 & \textbf{11.95} & \underline{11.64} & 10.04 & \textbf{14.66} & 13.82 & 13.26 & \textbf{14.66} \\
Negation                       & 5.96 & 6.72  & \textbf{8.51}  & \underline{7.48}  & 6.97 & \textbf{9.91}  & 8.84  & \underline{9.42}  & 8.82  \\
Direct Addressing             & 5.55 & 11.53 & \underline{12.41} & \textbf{12.94} & 10.59 & 15.18 & 14.84 & \underline{15.29} & \textbf{16.02} \\
Compare \& Change                & 4.30  & \textbf{8.09}  & 8.02  & \underline{8.08}  & 7.48  & \textbf{9.52}  & \underline{9.48}  & 8.30   & 9.42  \\
Comparative Statement         & 5.82 & 8.38  & \textbf{10.16} & \underline{9.81}  & 8.47  & \textbf{12.10}  & 11.19 & 10.27 & \underline{11.60}  \\
Statement w/ Conjunction   & 5.25 & 9.35  & \underline{10.46} & \textbf{10.54} & 8.94  & \underline{13.20}  & 12.73 & 12.76 & \textbf{13.47} \\
Spatial Rel. \& Background & 5.89 & 11.30  & \underline{11.74} & \textbf{12.48} & 9.97  & \underline{15.24} & 14.18 & 14.25 & \textbf{15.75} \\
Viewpoint                      & 4.07 & 7.42  & \textbf{7.98}  & \underline{7.45}  & 4.52  & 7.86  & \textbf{8.51}  & 8.14  & \underline{8.25}  \\
  \bottomrule
  \end{tabular}}
  \label{tab:circo_semantic_aspects}
\end{table*}
\tit{CIRCO}
In \Cref{tab:circo_test}, we report the results for the CIRCO test set. Firstly, we observe that, in contrast to FashionIQ and CIRR, Image+Text outperforms Image-only and Text-only. This result indicates that CIRCO contains queries where both the reference image and the relative caption are equally crucial for retrieving the target images. Secondly, \methodpami achieves a considerable improvement over all the baselines and even outperforms Pic2Word, despite a smaller backbone. Considering the L/14 backbone, \methodpami-XL would achieve the best results. However, we recall that the conference version of our method \methodiccv-XL \cite{baldrati2023zero} was used to ease the annotation process of CIRCO. Therefore, since the core of the approach proposed in this work is the same, it is likely that the results could exhibit some sort of bias. Still, we report them for completeness.

\Cref{tab:circo_semantic_aspects} shows the $mAP@10$ results on the CIRCO test set for each semantic category. We observe that some semantic aspects, such as \textit{viewpoint} and \textit{negation}, pose a significant challenge for all the reported methods. We suppose this outcome is due to the use of CLIP, which struggles to comprehend specific language constructs, such as negations and compositional relationships between objects and attributes \cite{wang2022learn, yuksekgonul2022and}. On the other hand, all the considered methods seem to handle semantic categories such as \textit{addition} and \textit{direct addressing} more effectively. We argue that this result stems from the fact that the relative captions corresponding to these semantic aspects have a structure similar to that of the absolute captions employed for pre-training CLIP. Regarding the comparison between different approaches, the considerations we made for \cref{tab:circo_test} still apply, with the proposed method achieving state-of-the-art performance across different semantic aspects.

Thanks to the semantic annotation phase introduced in this work, CIRCO allows such a fine-grained analysis of the results. Consequently, it is possible to discern the intricate complexities inherent in different query types, thereby guiding targeted research efforts toward tackling these challenges.

\tit{Domain Conversion}
\Cref{tab:imagenet_pic2word_test} illustrates the results for the domain conversion task. Following \cite{saito2023pic2word}, the query images are sourced from 200 classes of ImageNet \cite{russakovsky2015imagenet} validation set, while the target ones belong to ImageNet-R \cite{hendrycks2021many}. We recall that we relied on the ImageNet unlabeled test set as the pre-training dataset of our method, so there is no overlap with the evaluation images of the domain conversion task. The purpose of this experiment is to study how our model can convert the domain of a query image by using the prompt ``$\{domain\}$ of $S_{*}$", where $\{domain\}$ is a word that indicates the domain, \eg \textit{toy} or \textit{origami}. We consider the retrieved image correct if its class is the same as that of the query image and its domain matches the one specified by the prompt. For instance, given the domain ``toy" and a query image containing a real-world shark, the goal is to retrieve images depicting a shark toy. The results show that, both for the B/32 and L/14 backbones, our method outperforms all the baselines. 

\begin{table*}[]
  \centering
  \Large
  \caption{Quantitative results for the domain conversion task. The query images are from the ImageNet validation set, while the target ones belong to ImageNet-R. $^{\dagger}$ indicates results from the original paper.}

  \resizebox{0.81\linewidth}{!}{ 
  \begin{tabular}{clcccccccccc} 
  \toprule
  \multicolumn{1}{c}{} & \multicolumn{1}{c}{} & \multicolumn{2}{c}{Cartoon} & \multicolumn{2}{c}{Origami} & \multicolumn{2}{c}{Toy} & \multicolumn{2}{c}{Sculpture} & \multicolumn{2}{c}{Average} \\
  \cmidrule(lr){3-4}
  \cmidrule(lr){5-6}
  \cmidrule(lr){7-8}
  \cmidrule(lr){9-10}
  \cmidrule(lr){11-12}
  \multicolumn{1}{c}{Backbone} & \multicolumn{1}{l}{Method} & R$@10$ & R$@50$ & R$@10$ & R$@50$ & R$@10$ & R$@50$ & R$@10$ & R$@50$ & R$@10$ & R$@50$ \\ 
  \midrule
  \multirow{9}{*}{B/32} & Image-only & 0.22 & 3.09 & 0.43 & 2.38 & 0.55 & 4.53 & 0.47 & 3.80 & 0.42 & 3.45 \\
  & Text-only & 0.16 & 1.14 & 1.17 & 5.29 & 0.31 & 1.14 & 0.31 & 1.83 & 0.49 & 2.35 \\ 
  & Image + Text & 1.50 & 8.81 & 1.66 & 7.05 & 1.00 & 7.43 & 1.25 & 7.68 & 1.35 & 7.74 \\
  & Captioning & 6.75 & 18.58 & 9.60 & \underline{21.22} & 5.95 & 17.23 & 7.18 & 18.16 & 7.37 & 18.80 \\
  & PALAVRA \cite{cohen2022this} & 2.56 & 10.81 & 3.29 & 11.39 & 1.48 & 9.44 & 2.89 & 12.50 & 2.56 & 11.04 \\
  & \methodiccv-OTI \cite{baldrati2023zero} & 7.10 & 18.97 & 8.91 & 19.88 & 5.37 & 17.08 & 6.81 & 18.00 & 7.05 & 18.48 \\
  & \methodiccv \cite{baldrati2023zero} & 6.12 & 20.24 & 7.91 & 20.03 & 3.10 & 16.18 & 4.56 & 17.35 & 5.42 & 18.45 \\ 
  \rowcolor{LightCyan} \cellcolor{white} & \textbf{\methodpami-OTI} & \underline{9.49} & \underline{24.10} & \textbf{9.93} & \textbf{21.27} & \underline{6.96} & \underline{21.43} & \textbf{9.21} & \underline{22.50} & \underline{8.90} & \underline{22.33} \\
  \rowcolor{LightCyan} \cellcolor{white} & \textbf{\methodpami} & \textbf{10.02} & \textbf{25.01} & \underline{9.77} & 21.13 & \textbf{7.07} & \textbf{22.97} & \underline{9.16} & \textbf{22.93} & \textbf{9.01} & \textbf{23.01} \\
  \midrule[.02em]
  \multirow{7}{*}{L/14} & Pic2Word$^{\dagger}$ \cite{saito2023pic2word} & 8.00 & 21.90 & 13.50 & 25.60 & 8.70 & 21.60 & 10.00 & 23.80 & 10.10 & 23.20 \\
  & Context-I2W$^{\dagger}$ \cite{tang2023context} & 10.20 & 26.10 & 17.50 & 28.70 & \underline{11.60} & 27.40 & 12.10 & 28.20 & 12.90 & 27.60 \\
  & LinCIR \cite{gu2023language} & \underline{11.34} & 28.96 & 17.15 & 30.31 & \textbf{13.40} & \underline{30.30} & 13.19 & 28.43 & \underline{13.77} & 29.50 \\
  & \methodiccv-XL-OTI \cite{baldrati2023zero} & 9.85 & 24.97 & 18.81 & 30.55 & 10.19 & 27.26 & 12.75 & 28.94 & 12.90 & 27.93 \\
  & \methodiccv-XL \cite{baldrati2023zero} & 9.67 & 29.94 & 19.48 & 34.12 & 7.45 & 26.75 & 11.57 & 33.31 & 12.04 & 31.03 \\ 
  \rowcolor{LightCyan} \cellcolor{white} & \textbf{\methodpami-XL-OTI} & 10.48 & \underline{29.76} & \underline{20.01} & \underline{33.36} & 9.68 & 30.28 & \underline{13.39} & \underline{33.72} & 13.39 & \underline{31.78} \\
  \rowcolor{LightCyan} \cellcolor{white} & \textbf{\methodpami-XL} & \textbf{12.84} & \textbf{31.67} & \textbf{22.17} & \textbf{34.51} & 11.20 & \textbf{31.87} & \textbf{15.88} & \textbf{35.49} & \textbf{15.52} & \textbf{33.39} \\
  \bottomrule
  \end{tabular}}
  \label{tab:imagenet_pic2word_test}
\end{table*}

\tit{Object Composition}  
We provide the results for the object composition task on the COCO validation set \cite{lin2014microsoft} in \cref{tab:coco_pic2word_test}. This task aims to retrieve an image comprising an object specified with a single query image and other objects described with text. The object composition task closely resembles the personalized retrieval one \cite{cohen2022this}, differing mainly in that the latter involves queries composed of multiple images depicting the same object instance. Following \cite{saito2023pic2word}, we use the prompt ``a photo of $S_{*}$, $\{obj_{1}\}$ and $\{obj_{2}\}$, \ldots , and $\{obj_{n}\}$", where $\{obj_{i}\}$ are text descriptions of objects, \eg \textit{mouse}, \textit{laptop} or \textit{kite}. For both the B/32 and L/14 backbones, we notice that our approach achieves state-of-the-art results. Considering the L/14 backbone, we observe a performance gap between the OTI and $\phi$ variants of the proposed method. We suppose this result is due to the similarity between the object composition and personalized retrieval tasks, where it has been shown that an optimization-based method is more suitable and achieves better performance \cite{cohen2022this}. The results obtained by PALAVRA \cite{cohen2022this} on the object composition task confirm our hypothesis, as it achieves significantly better relative performance than in composed image retrieval, \eg on CIRCO. In addition, we observe that LinCIR \cite{gu2023language} obtains considerably worse performance than \methodpami-XL. We suppose this outcome is due to their language-only training strategy, which makes their model struggle to capture fine-grained visual details.

\begin{table}[]
  \centering
  \caption{Quantitative results for the object composition task on COCO validation set. $^{\dagger}$ indicates results from the original paper.}
  \large
  \resizebox{\linewidth}{!}{ 
  \begin{tabular}{clccc} 
  \toprule
  Backbone & Method & R$@1$ & R$@5$ & R$@10$ \\ 
  \midrule
  \multirow{9}{*}{B/32} & Image-only & 7.30 & 13.64 & 17.19 \\
  & Text-only & 4.93 & 14.06 & 21.51 \\ 
  & Image + Text & 8.96 & 18.24 & 23.92 \\
  & Captioning & 5.96 & 15.84 & 22.92 \\
  & PALAVRA \cite{cohen2022this} & \underline{12.94} & 25.66 & 32.40 \\
  & \methodiccv-OTI \cite{baldrati2023zero} & \textbf{13.03} & 26.00 & 34.27 \\
  & \methodiccv \cite{baldrati2023zero} & 10.91 & 24.58 & 33.28 \\
  \rowcolor{LightCyan} \cellcolor{white} & \textbf{\methodpami-OTI} & 11.96 & \textbf{26.63} & \underline{35.43} \\
  \rowcolor{LightCyan} \cellcolor{white} & \textbf{\methodpami} & 11.75 & \underline{26.40} & \textbf{35.87} \\
  \midrule[.02em]
  \multirow{7}{*}{L/14} & Pic2Word$^{\dagger}$ \cite{saito2023pic2word} & 11.50 & 24.80 & 33.40 \\
  & Context-I2W$^{\dagger}$ \cite{tang2023context} & 13.50 & 28.50 & 38.10 \\
  & LinCIR \cite{gu2023language} & 10.93 & 24.83 & 34.48 \\
  & \methodiccv-XL-OTI \cite{baldrati2023zero} & \underline{17.04} & \underline{31.43} & \underline{40.81} \\
  & \methodiccv-XL \cite{baldrati2023zero} & 14.21 & 29.02 & 37.71 \\
  \rowcolor{LightCyan} \cellcolor{white} & \textbf{\methodpami-XL-OTI} & \textbf{17.54} & \textbf{32.55} & \textbf{41.22} \\
  \rowcolor{LightCyan} \cellcolor{white} & \textbf{\methodpami-XL} & 15.01 & 30.05 & 38.76 \\
  \bottomrule
  \end{tabular}}
  \label{tab:coco_pic2word_test}
\end{table}

\tit{Discussion} 
Our experiments show that the proposed approach consistently achieves commendable performance across different datasets, highlighting its robustness. Compared to the baselines, our method has better generalization and adaptability capabilities to tasks beyond standard composed image retrieval, such as the domain conversion and object composition. Moreover, \methodpami and \methodpami-OTI obtain comparable results in most scenarios. This confirms the effectiveness of the distillation process, which offers a significant efficiency improvement without sacrificing the performance. Finally, the fine-grained evaluation on CIRCO reveals the intrinsic limitations of current ZS-CIR methods, which struggle when dealing with queries involving complex semantic aspects such as negation and viewpoint changes.

\subsection{Ablation Studies}
We conduct extensive ablation studies to measure the individual contribution of each component of our approach. To avoid potential interferences, we evaluate the two main stages of the proposed method separately. In particular, we assess the performance of the textual inversion network $\phi$ while keeping fixed the collection of OTI pre-generated tokens obtained as detailed in \cref{sec:optimization}. As $\phi$ distills the knowledge of the OTI pre-generated tokens, we assume that the more informative they are (\ie the better OTI performs), the better the performance achieved by $\phi$ will be. We rely on the CIRR and FashionIQ validation sets to conduct the ablation studies and report the results for the main evaluation metrics. Specifically, for FashionIQ we report the average scores. We focus solely on the B/32 version of our method for simplicity.

\tit{Optimization-based textual inversion (OTI)}\label{sec:ablation_oti}
We ablate each of the components of the optimization process: 1) \textit{w/o GPT reg}: we regularize with a prompt containing only the concept word, without the GPT-generated suffix; 2) \textit{random reg}: we additionally substitute the concept word with a random word; 3) \textit{w/o reg}: we completely remove the regularization loss; 4) \textit{w/o noise}: we do not add Gaussian noise to the text features, \ie we set $\gamma$ to 0; 5) \textit{$L_{2}$ loss}: we substitute the cosine loss in \cref{eq:loss_content,eq:loss_oti_gpt} with an $L_{2}$-based one.

The upper part of \cref{tab:ablation} shows the results. As a different loss leads to a different speed of convergence, we use a tailored number of optimization iterations for each ablation experiment and report the best performance. First, we find that regularization plays a crucial role in ensuring that the pseudo-word tokens reside in the CLIP token embedding manifold and can effectively interact with the CLIP vocabulary tokens. In particular, we argue that our GPT-based regularization loss allows the pseudo-word tokens to interact with text resembling human-written language, thereby improving their communication with the relative captions and ultimately enhancing retrieval performance. This effect is particularly pronounced in CIRR, where relative captions tend to be more elaborate and have a more diverse vocabulary. Then, we observe that using a cosine loss obtains better performance than an $L_{2}$ one. We suppose that this outcome stems from the CLIP training strategy, which uses a cosine similarity-based loss. Finally, we notice that adding Gaussian noise to the text features in the $\mathcal{L}_{content}$ loss computation improves the performance. This result shows that our strategy for mitigating the effect of the modality gap is fruitful.

\tit{Textual inversion network $\vect{\phi}$} \label{sec:ablation_phi}
We ablate the losses we use during the pre-training of $\phi$: 1) \textit{cos distil}: we use a cosine distillation loss instead of a contrastive one;  2) \textit{w/o distil}: we replace $\mathcal{L}_{distil}$ with the cycle contrastive loss introduced by \cite{cohen2022this}, which directly considers the image and text features; 3) \textit{w/o reg}: we remove the $\mathcal{L}_{gpt}$ regularization loss; 4) \textit{w/o $\mathcal{L}_{pen}$}: we remove the regularization penalty term on the predicted pseudo-word tokens, \ie we set $\lambda_{pen}$ to 0; 5) \textit{w/o HNSS}: we compose batches randomly instead of using the hard negative sampling strategy described in \cref{sec:phi_training}, \ie we set $\alpha$ to 0.

We report the results in the lower section of \cref{tab:ablation}. The contrastive version of the distillation loss achieves better performance than the cosine one. Compared to the cycle contrastive loss, our distillation-based loss proves to be significantly more effective, highlighting how learning from OTI pre-generated tokens is more fruitful than learning from raw images. Moreover, although the pre-generated pseudo-word tokens are already regularized, we observe that our GPT-based regularization loss is still beneficial for training $\phi$, especially on the CIRR dataset. Regarding $\mathcal{L}_{pen}$, we find that introducing an additional regularization penalty term to constrain the predicted pseudo-word tokens is effective, as it helps in making them reside in the CLIP manifold \cite{gal2023encoder}. Finally, we observe that by removing the proposed hard negative sampling strategy we achieve comparable performance on CIRR but worse on FashionIQ. This outcome is due to the narrow domain of FashionIQ, as images are closely related and subtle details are crucial. This confirms that our hard negative sampling strategy helps the model in capturing fine-grained details.

\tit{Hyperparameters}
We study the effect of the main hyperparameters of our method and report the results in \cref{tab:ablation_hyperparameter}. Regarding the standard deviation $\gamma$ of the noise, we observe a sweet spot ($\gamma = 0.64$) that balances mitigating the modality gap and preserving the informative content of the text features.
For the ratio $\alpha$ of hard negative samples per batch, composing batches entirely of hard negatives ($\alpha=1$) leads the model to focus on fine-grained details, improving performance on fine-grained datasets like FashionIQ. However, this comes at the cost of reduced image diversity within each batch, resulting in worse performance on broader-domain datasets such as CIRR. A moderate value ($\alpha=0.5$) achieves a better trade-off by incorporating challenging examples while maintaining a diverse range of image content.

\begin{table}[!t]
  \centering
  \caption{Ablation studies on CIRR and FashionIQ validation sets. For FashionIQ, we consider the average recall. } 
  \large
  \resizebox{\linewidth}{!}{ 
  \begin{tabular}{clccccc} 
  \toprule
  \multicolumn{1}{c}{} & \multicolumn{1}{c}{} & \multicolumn{2}{c}{FashionIQ} & \multicolumn{3}{c}{CIRR}\\
  \cmidrule(lr){3-4}
  \cmidrule(lr){5-7}
  \multicolumn{1}{l}{Abl.} & \multicolumn{1}{l}{Method} & R$@10$ & R$@50$ & R$@1$ & R$@5$ & R$@10$ \\ \midrule
   \multirow{6}{*}{OTI} & w/o GPT reg & 21.63 & 41.40 & 21.04 & 50.99 & 64.21 \\
  & random reg & 21.53 & 40.01 & 21.21 & 48.43 & 62.97 \\
  & w/o reg & 19.29 & 37.10 & 17.43 & 46.85 & 60.65 \\
  & w/o noise & 23.62 & 43.80 & 24.75 & 55.39 & 69.28 \\
  & $L_{2}$ loss & \underline{24.40} & \underline{44.71} & \underline{25.23} & \underline{56.80} & \textbf{70.46} \\
  \rowcolor{LightCyan} \cellcolor{white} & \textbf{\methodpami-OTI} & \textbf{25.06} & \textbf{44.79} & \textbf{25.57} & \textbf{57.11} & \textbf{70.46} \\ \midrule[.02em]
  \multirow{6}{*}{$\phi$} & cos distil & 22.46 & 42.26 & 24.80 & 54.36 & 68.00 \\
  & w/o distil & 19.64 & 38.54 & 21.93 & 49.94 & 63.55 \\
  & w/o reg & \underline{24.33} & \textbf{45.08} & 24.63 & 56.20 & 69.22 \\
  & w/o $\mathcal{L}_{pen}$ & 24.19 & 44.70 & 25.66 & 57.14 & 70.22 \\
  & w/o HNSS & 23.63 & 44.46 & \underline{25.70} & \textbf{57.45} & \underline{70.24} \\
  \rowcolor{LightCyan} \cellcolor{white} & \textbf{\methodpami} & \textbf{24.40} & \underline{44.80} & \textbf{25.74} & \underline{57.35} & \textbf{70.32} \\
  \bottomrule
  \end{tabular}}
  \label{tab:ablation}
\end{table}

\begin{table}[!t]
  \centering
  \caption{Ablation studies on the choice of hyperparameters on CIRR and FashionIQ validation sets. For FashionIQ, we consider the average recall.} 
  \resizebox{\linewidth}{!}{ 
  \begin{tabular}{clccccc} 
  \toprule
  \multicolumn{1}{c}{} & \multicolumn{1}{c}{} & \multicolumn{2}{c}{FashionIQ} & \multicolumn{3}{c}{CIRR}\\
  \cmidrule(lr){3-4}
  \cmidrule(lr){5-7}
  \multicolumn{1}{l}{Abl.} & \multicolumn{1}{l}{Value} & R$@10$ & R$@50$ & R$@1$ & R$@5$ & R$@10$ \\ \midrule
    & $\gamma = 0.32$ & \underline{24.76} & \underline{44.73} & 25.16 & \underline{56.92} & 69.98 \\
  \rowcolor{LightCyan} \cellcolor{white} & \textbf{$\gamma = 0.64$} & \textbf{25.06} & \textbf{44.79} & \underline{25.57} & \textbf{57.11} & \textbf{70.46} \\
  \multirow{-3}{*}{OTI} & $\gamma = 0.96$ & 24.15 & 44.57 & \textbf{25.60} & 56.54 & \textbf{70.46} \\ \midrule[.02em]
  & $\alpha = 0.25$ & 23.76 & 44.49 & \underline{25.71} & \textbf{57.42} & \underline{70.28} \\
  \rowcolor{LightCyan} \cellcolor{white} & \textbf{$\alpha = 0.5$} & \underline{24.40} & \underline{44.80} & \textbf{25.74} & \underline{57.35} & \textbf{70.32} \\
   \multirow{-3}{*}{$\phi$} & $\alpha = 1$ & \textbf{24.52} & \textbf{44.95} & 25.14 & 56.83 & 70.06 \\
  \bottomrule
  \end{tabular}}
  \label{tab:ablation_hyperparameter}
\end{table}

\begin{table}[!t]
  \centering
  \caption{Evaluation of the effect of the $\phi$ pre-training dataset on the respective evaluation split of FashionIQ, CIRR, and CIRCO. For FashionIQ, we consider the average recall.} 
  \Large
  \resizebox{\linewidth}{!}{ 
  \begin{tabular}{lcccccc} 
  \toprule
  \multicolumn{1}{c}{} & \multicolumn{2}{c}{FashionIQ} & \multicolumn{3}{c}{CIRR} & \multicolumn{1}{c}{CIRCO}\\
  \cmidrule(lr){2-3}
  \cmidrule(lr){4-6}
  \cmidrule(lr){7-7}
  \multicolumn{1}{l}{Method} & R$@10$ & R$@50$ & R$@1$ & R$@5$ & R$@10$ & mAP$@10$ \\ \midrule
  \methodpami-FIQ & \textbf{25.61} & \textbf{45.57} & 25.01 & 55.06 & 67.90 & 10.77 \\
  \methodpami-CIRR & 23.75 & 44.07 & 25.21 & 54.24 & \underline{68.15} & 10.49 \\
  \methodpami-NABirds & 23.27 & 42.81 & 24.43 & 53.81 & 66.96 & 8.92 \\
  \methodpami-FFHQ & 22.91 & 43.70 & 24.68 & 54.10 & 67.71 & 10.74 \\
  \methodpami-VGGFace2 & 23.20 & 43.72 & 24.96 & 54.75 & 67.83 & \underline{10.97} \\
  \rowcolor{LightCyan} \textbf{\methodpami-OTI} & \underline{25.06} & \underline{44.79} & \textbf{26.19} & \underline{55.18} & \textbf{68.55} & 10.94 \\
  \rowcolor{LightCyan} \textbf{\methodpami} & 24.40 & 44.80 & \underline{25.23} & \textbf{55.69} & 68.05 & \textbf{11.24} \\
  \bottomrule
  \end{tabular}}
  \label{tab:phi_different_datasets}
\end{table}

\subsection{Additional Experiments}\label{sec:additional_experiments}
\tit{Effect of $\vect{\phi}$ Pre-training Dataset}
We carry out several experiments to study the impact of the $\phi$ pre-training dataset. Specifically, besides the version of $\phi$ trained on the test split of ImageNet1K, we also train some variants using: 1) CIRR training set, with 17K images; 2) FashionIQ training set, with 45K images; 3) NABirds \cite{van2015building} whole dataset, with 48K images depicting birds; 4) FFHQ \cite{karras2019style} training set, with 50K images of aligned and cropped faces; 5) a subset of VGGFace2 \cite{cao2018vggface2} we obtained by randomly sampling 5 images per subject, resulting in 45K images of faces. Regarding CIRR and FashionIQ, we use only the raw images without considering the associated labels to keep the approach unsupervised. By relying on these two datasets, we assess the impact of pre-training $\phi$ in a domain aligned with that of the testing dataset. Conversely, training on NABirds, FFHQ, or VGGFace2 provides insights into how effectively our method adapts to domains entirely distinct from the pre-training one. We selected these three datasets specifically for their very narrow domains, yet ensuring a sufficient number of images.

\Cref{tab:phi_different_datasets} reports the results. We notice how \methodpami-FIQ achieves the best performance on FashionIQ and even outperforms \methodpami-OTI, thus highlighting the effectiveness of our distillation-based approach. This result shows that pre-training on images belonging to the same domain as that of the testing ones leads to a performance gain. Moreover, \methodpami-FIQ also manages to generalize to a broader domain obtaining promising results on both CIRR and CIRCO. Despite relying only on 17K training images, \methodpami-CIRR achieves noteworthy results, suggesting that our approach is effective even in a low-data regime. Finally, we observe that \methodpami-FFHQ, \methodpami-NABirds, and \methodpami-VGGFace2 obtain promising results despite being pre-trained on datasets that have highly specific domains that are extremely different from those of the testing datasets. To contextualize, Pic2Word \cite{saito2023pic2word} scores a $R@1$ of 23.90 on CIRR and a $mAP@10$ of 9.51 on CIRCO. In comparison, \methodpami-VGGFace2, despite using a smaller backbone and being trained on such a narrow domain as human faces, achieves a $R@1$ of 24.96 on CIRR and a $mAP@10$ of 10.97 on CIRCO. These results show that our approach is robust to the $\phi$ pre-training dataset. Moreover, our model demonstrates noteworthy generalization capabilities, making it well-suited for application in any real-world scenario without the requirement for domain-specific pre-training.

 \begin{table}[!t]
  \centering
    \caption{Evaluation of the visual information embedded in $v_*$ for different regularization techniques on CIRR validation set. IR and CIR stand for Image Retrieval and Composed Image Retrieval, respectively.}
  \label{tab:vstar_image_retrieval}
  \Large
  \resizebox{\linewidth}{!}{ 
  \begin{tabular}{clccccccc} 
  \toprule
  \multicolumn{1}{c}{} & \multicolumn{1}{c}{} & \multicolumn{3}{c}{IR} & \multicolumn{1}{c}{} & \multicolumn{3}{c}{CIR}\\
  \cmidrule(lr){3-5}\cmidrule(lr){7-9}
  \multicolumn{1}{l}{Ablation} & \multicolumn{1}{l}{Method} & R$@1$ & R$@3$ & R$@5$ &  & R$@1$ & R$@5$ & R$@10$  \\ \midrule
    \multirow{4}{*}{OTI}  & w/o GPT reg & 99.63 & \textbf{100} & \textbf{100} && 21.04 & \underline{50.99} & \underline{64.21} \\
  & random reg  & 99.26 & 99.95 & 99.95 && \underline{21.21} & 48.43 & 62.97 \\
  & w/o reg  & \textbf{99.77} & \textbf{100} & \textbf{100} && 17.43 & 46.85 & 60.65\\
  \rowcolor{LightCyan} \cellcolor{white} & \textbf{\methodpami-OTI} & \textbf{99.77} & \textbf{100} & \textbf{100} && \textbf{25.57} & \textbf{57.11} & \textbf{70.46} \\ \midrule[.02em]
  \multirow{3}{*}{$\phi$}
  & w/o reg & \textbf{99.21} & \textbf{99.95} & \underline{99.95}  && 24.63 & 56.20 & 69.22 \\
  & w/o $\mathcal{L}_{pen}$ & \underline{98.98} & \textbf{99.95} & \textbf{100}  && \underline{25.66} & \underline{57.14} & \underline{70.22}   \\
  \rowcolor{LightCyan} \cellcolor{white} & \textbf{\methodpami}& 98.89 & \textbf{99.95} & \underline{99.95} &&  \textbf{25.74} & \textbf{57.35} & \textbf{70.32} \\
  \bottomrule
  \end{tabular}}

\end{table}
\tit{Visual Information in $\vect{v_*}$}
We carry out an image retrieval experiment by studying whether the pseudo-word tokens can retrieve the corresponding images. In this way, we assess the effectiveness of the pseudo-word tokens in capturing visual information. Starting from an image $I$, we obtain the corresponding pseudo-word token $v_*$ and its associated pseudo-word $S_*$ through textual inversion. We craft a generic prompt including the pseudo-word $S_*$, such as ``a photo of $S_*$". Then, we extract the text features via the CLIP text encoder and use them to query an image database. We expect the image $I$ to be the top-ranked result if the pseudo-word token manages to effectively embed its visual content.

\cref{tab:vstar_image_retrieval} shows the results for Image Retrieval (IR) alongside the corresponding ones for Composed Image Retrieval (CIR). We use the CIRR validation set to conduct the experiments. We report the results of all the ablation studies related to the regularization technique for both OTI and $\phi$. Refer to \cref{sec:ablation_oti} for more details on the setting of each ablation. Regardless of the regularization strategy, we observe that $v_*$ effectively captures the visual information of the image, achieving almost perfect IR scores. However, we obtain a significant performance improvement in CIR when relying on our GPT-powered loss. This highlights how our regularization technique enhances the ability of the pseudo-word tokens to interact with the actual words composing the relative caption while preserving the visual information embedded in $v_*$.

\tit{Comparison with Supervised Baselines}
\begin{table}[!t]
  \centering
  \caption{Comparison with supervised baselines on CIRR and FashionIQ validation sets. For FashionIQ, we consider the average recall.}
  \resizebox{\linewidth}{!}{ 
  \begin{tabular}{lccccc} 
  \toprule
  \multicolumn{1}{c}{} & \multicolumn{2}{c}{FashionIQ} & \multicolumn{3}{c}{CIRR} \\
  \cmidrule(lr){2-3}
  \cmidrule{4-6}
  \multicolumn{1}{l}{Method} & R$@10$ & R$@50$ & R$@1$ & R$@5$ & R$@10$ \\ \midrule
  Combiner-FIQ~\cite{baldrati2022effective} & \textbf{32.96} & \textbf{54.55} & 19.88 & 48.05 & 61.11 \\
  Combiner-CIRR~\cite{baldrati2022effective} & 20.91 & 40.40 & \textbf{32.24} & \textbf{65.46} & \textbf{78.21} \\
  \rowcolor{LightCyan} \textbf{\methodpami-OTI} & \underline{25.06} & 44.79 & 25.57 & 57.11 & \underline{70.46} \\
  \rowcolor{LightCyan} \textbf{\methodpami} & 24.40 & \underline{44.80} & \underline{25.74} & \underline{57.35} & 70.32 \\  
  \bottomrule
  \end{tabular}}
  \label{tab:supervised_baselines}
\end{table}
We measure the generalization capabilities of supervised CIR models by performing a comparison with our zero-shot approach. In particular, we consider Combiner \cite{baldrati2022effective}, which fuses image and text CLIP features through a combiner network. We chose Combiner as we believe it represents the most similar method to ours among the supervised ones, as we both rely on an out-of-the-box CLIP model. We train two versions of Combiner based on the B/32 backbone on the FashionIQ and CIRR training sets, respectively, using the official repository. We test both Combiner versions on FashionIQ and CIRR validation sets and report the results in \cref{tab:supervised_baselines}. As expected, Combiner achieves the best performance when the training and testing datasets correspond. However, both supervised models struggle to generalize to different domains, as also observed by \cite{saito2023pic2word}. Conversely, \methodpami exhibits remarkable performance on both datasets in a zero-shot manner. Thus, given that we do not require a costly manually annotated training set, the proposed method demonstrates better scalability and suitability for real-world applications of composed image retrieval.

\section{Conclusion} \label{sec:conclusion}
In this work we expand upon our conference paper and introduce a new task, Zero-Shot Composed Image Retrieval (ZS-CIR), aimed at tackling CIR without requiring an expensive labeled training dataset. Since its introduction, several works have addressed ZS-CIR, highlighting its significance and relevance to the research community. We present an approach, named \methodpami, that involves pre-training a lightweight textual inversion network via a distillation loss to retain the expressiveness of an optimization-based method while achieving a substantial efficiency gain. In addition, we introduce an open-domain benchmarking dataset for CIR, named CIRCO. CIRCO is the first CIR dataset featuring multiple labeled ground truths, reduced false negatives, and a semantic categorization of the queries. \methodpami achieves state-of-the-art performance on FashionIQ, CIRR and the proposed CIRCO. Moreover, the proposed approach demonstrates better generalization capabilities than competing methods, as shown by two additional evaluation settings, namely object composition and domain conversion.

\section*{Acknowledgments}
This work was partially supported by the European Commission under European Horizon 2020 Programme, grant number 101004545 - ReInHerit.

\bibliographystyle{IEEEtran}
\bibliography{bib}

\begin{IEEEbiography}[{\includegraphics[width=1in,height=1.25in,clip,keepaspectratio]{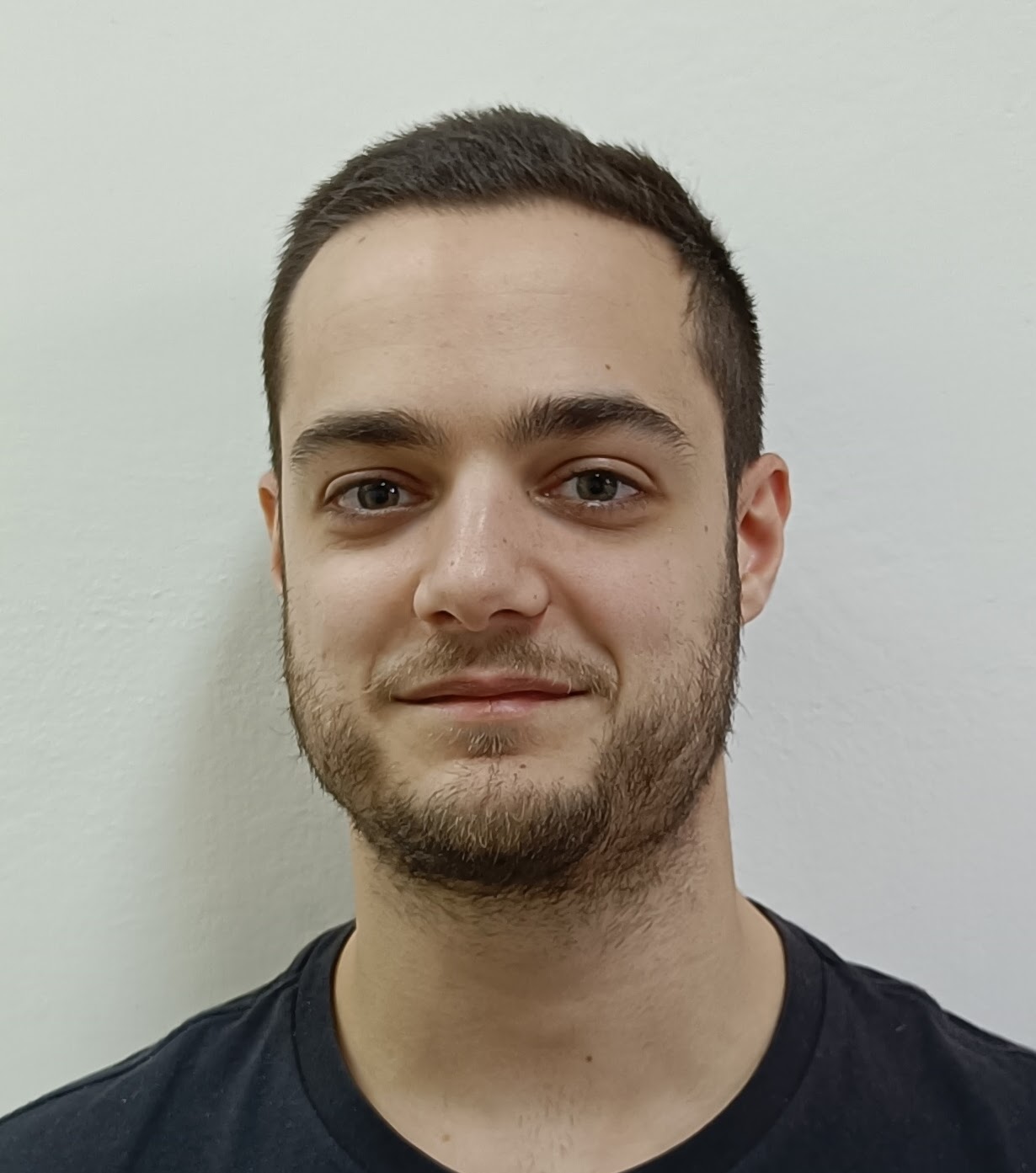}}]{Lorenzo Agnolucci}
received the M.S.~degree (cum laude) in Computer Engineering from the University of Florence, Italy, in 2021.
Currently, he is a PhD student at the University of Florence at the Media Integration and Communication Center (MICC). His research interests revolve around machine learning and computer vision, with a particular focus on low-level vision and vision-language models.\end{IEEEbiography}

\begin{IEEEbiography}[{\includegraphics[width=1in,height=1.25in,clip,keepaspectratio]{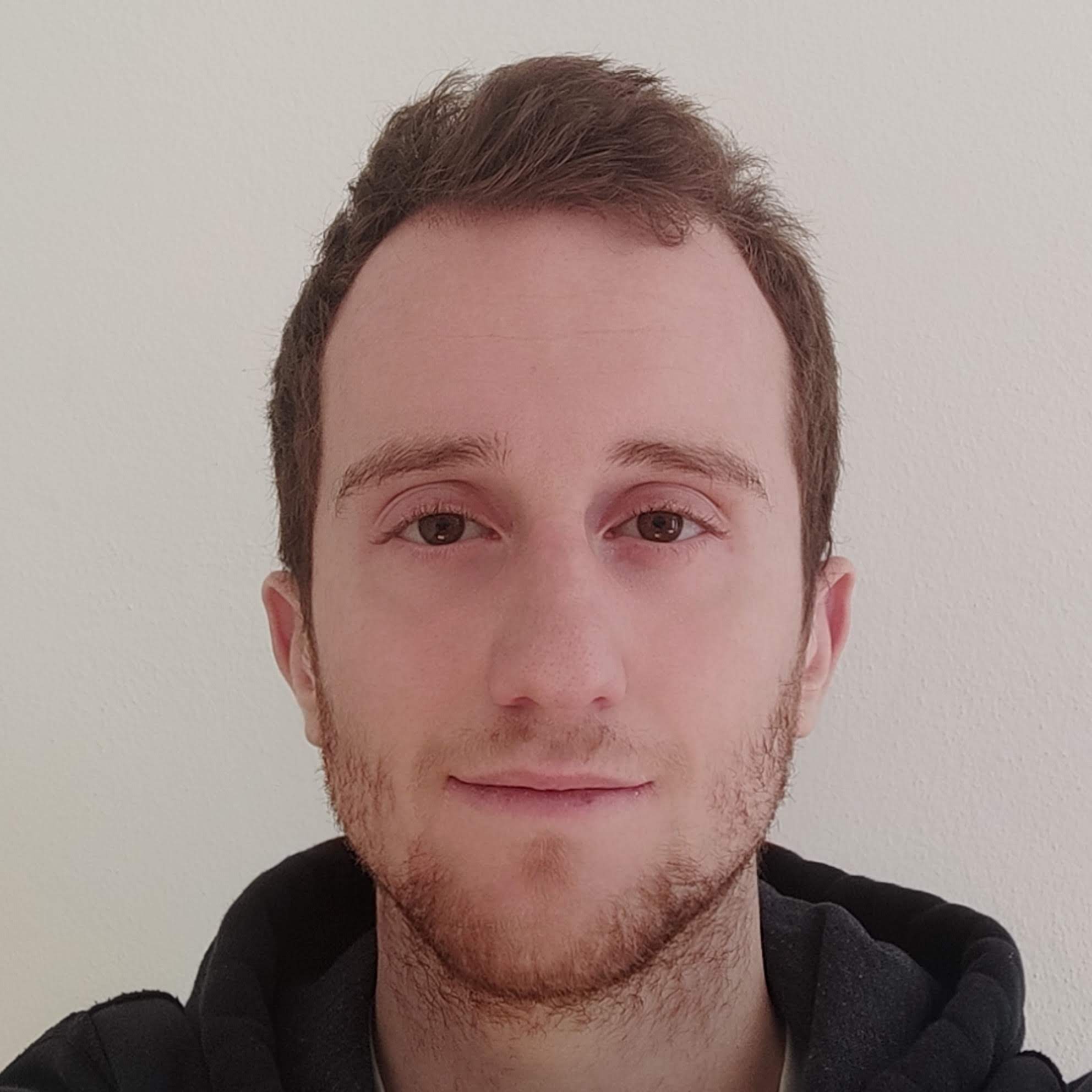}}]{Alberto Baldrati} received the M.S.~degree (cum laude) in Computer Engineering from the University of Florence, Italy, in 2021.
Currently, he is a Ph.D. student enrolled in the Italian National PhD Program in AI at the University of Pisa, while actively conducting research at the Media Integration and Communication Center (MICC) affiliated with the University of Florence. His research interests include machine learning and computer vision, focusing on vision and language, composed image retrieval, and fashion image generation.
\end{IEEEbiography}

\begin{IEEEbiography}[{\includegraphics[width=1in,height=1.25in,clip,keepaspectratio]{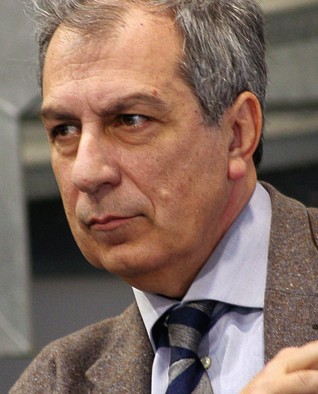}}]{Alberto Del Bimbo}
(Senior Member, IEEE) received the master’s degree cum laude in electrical engineering, in 1977.
He is a Full Professor of computer engineering, and the Director of the Media Integration and Communication Center, University of Florence, Florence, Italy. From 1996 to 2000, he was the President of the IAPR Italian Chapter and from 1998 to 2000, the Member-at- Large with the IEEE Publication Board. His research interests include multimedia information
retrieval, pattern recognition, and computer vision.
Prof. Del Bimbo received the SIGMM Technical Achievement Award for Outstanding Technical Contributions to Multimedia Computing, Communications and Applications. He was nominated the ACM Distinguished Scientist in 2016. He is a co-founder of Small Pixels, an academic spin-off working on visual quality improvement based on AI.\end{IEEEbiography}

\begin{IEEEbiography}[{\includegraphics[width=1in,height=1.25in,clip,keepaspectratio]{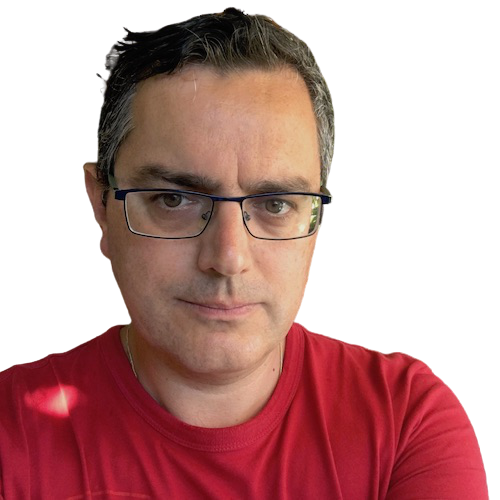}}]{Marco Bertini}
is an associate professor of computer science at the School of Engineering of the University of Florence and director of the Media Integration and Communication Center (MICC) at the same university. His interests regard computer vision, multimedia, pattern recognition, and their application to different domains such as cultural heritage. He has been general co-chair, program co-chair and area chair of several international conferences and workshops on multimedia and computer vision (ACM MM, ICMR, CBMI, etc.), and was associate editor of IEEE Transactions on Multimedia.
He has been involved in different roles in more than 10 EU research projects. He is a co-founder of Small Pixels, an academic spin-off working on visual quality improvement based on AI.\end{IEEEbiography}

\vfill

\end{document}